\documentclass[a4paper,fleqn]{cas-dc}
\usepackage{graphicx}
\usepackage{subfigure}

\usepackage{caption2}

\usepackage[numbers,sort&compress]{natbib}

\usepackage[T1]{fontenc}
\usepackage{algorithm}
\usepackage{algorithmic}

\def\tsc#1{\csdef{#1}{\textsc{\lowercase{#1}}\xspace}}
\tsc{WGM}
\tsc{QE}

\begin{document}
\let\WriteBookmarks\relax
\def\floatpagepagefraction{1}
\def\textpagefraction{.001}

\shorttitle{Deep learning-based person re-identification methods: A survey and outlook of recent works}  
\shortauthors{Zhangqiang Ming}  

\title [mode = title]{Deep learning-based person re-identification methods: A survey and outlook of recent works}  



%

\author{Zhangqiang Ming}[orcid=0000-0003-1616-8054]



\ead{mingzhangqiang@stu.scu.edu.cn}


\credit{Conceptualization, Methodology, Investigation, Writing-original draft, Revision}

\affiliation{organization={College of Computer Science},
	addressline={Sichuan University}, 
	city={Chengdu},
	postcode={610065}, 
	country={China}}

\author{Min Zhu}[orcid=0000-0002-5664-1558]
\cormark[1]
\ead{zhumin@scu.edu.cn}
\credit{onceptualization, Writing - Review \& Editing, Supervision, Resources}
\cortext[1]{Corresponding author}

\author{Xiangkun Wang}
\credit{Writing-Review, Funding Acquisition}

\author{Jiamin Zhu}
\credit{Data curation, Visualization, Investigation}

\author{Junlong Cheng}
\credit{Writing-Review, Data curation}

\author{Chengrui Gao}
\credit{Writing-Review, Investigation}

\author{Yong Yang}
\credit{Writing-Review, Data curation}

\author{Xiaoyong Wei}
\credit{Data Curation, Visualization, Validation}

\begin{abstract}
In recent years, with the increasing demand for public safety and the rapid development of intelligent surveillance networks, person re-identification (Re-ID) has become one of the hot research topics in the computer vision field. The main research goal of person Re-ID is to retrieve persons with the same identity from different cameras. However, traditional person Re-ID methods require manual marking of person targets, which consumes a lot of labor cost. With the widespread application of deep neural networks, many deep learning-based person Re-ID methods have emerged. Therefore, this paper is to facilitate researchers to understand the latest research results and the future trends in the field. Firstly, we summarize the studies of several recently published person Re-ID surveys and complement the latest research methods to systematically classify deep learning-based person Re-ID methods. Secondly, we propose a multi-dimensional taxonomy that classifies current deep learning-based person Re-ID methods into four categories according to metric and representation learning, including methods for deep metric learning, local feature learning, generative adversarial learning and sequence feature learning. Furthermore, we subdivide the above four categories according to their methodologies and motivations, discussing the advantages and limitations of part subcategories. Finally, we discuss some challenges and possible research directions for person Re-ID.
\end{abstract}


\begin{highlights}
\item The main contributions of person Re-ID surveys are summarized and discussed in recent years.

\item A metric and representation learning-based taxonomy is provided for recent person Re-ID methods.

\item The above main categories are subdivided based on their methodologies and motivations.

\item The advantages and limitations of part subcategories are summarized.

\item Furthermore, some challenges and possible research directions for person Re-ID are discussed.
\end{highlights}


\maketitle
\section{Introduction}\label{Introduction} 
\begin{figure}[tbh]
	\centering
	\includegraphics[width=0.95\linewidth]{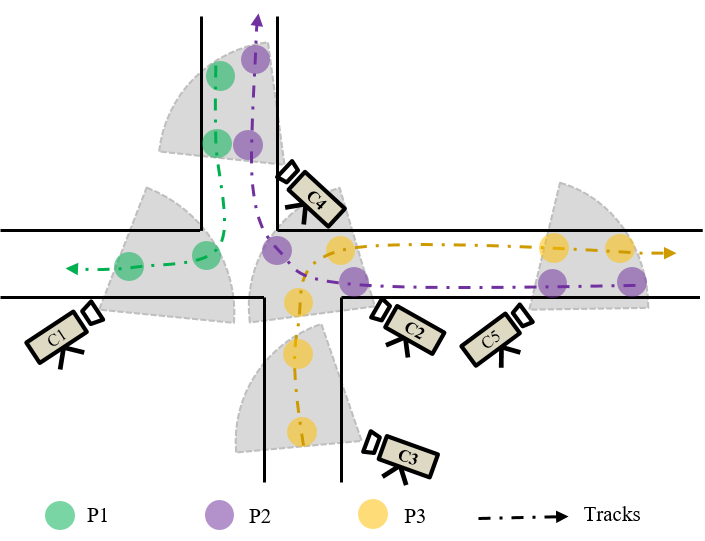}
	\caption{Multi-camera surveillance network illustration of person Re-ID.} 
	\label{fig:1-1-Multi-camera-Re-ID}
\end{figure}

In recent years, with the rapid development of intelligent surveillance devices and the increasing demand for public safety, a large number of cameras have been deployed in public places such as airports, communities, streets and campuses. These camera networks typically span large geographic areas with non-overlapping coverage and generate a large amount of surveillance video every day. 
We use this video data to analyze the activity patterns and behavioral characteristics of pedestrians in the real world for applications such as target detection, multi-camera target tracking and crowd behavior analysis. Person Re-ID can be traced back to the problem of multi-target multi-camera tracking (MTMCT tracking) \citep{Wang2013IntelligentMulti}, which aims to determine whether pedestrians captured by different cameras or pedestrian images from different video clips of the same camera are the same pedestrian \citep{Luo2019SurveyDeep}. \autoref{fig:1-1-Multi-camera-Re-ID} illustrates an example of a surveillance area monitored by multiple cameras with non-overlapping fields of view.

\begin{figure*}[tbh]
	\centering
	\includegraphics[width=0.95\linewidth]{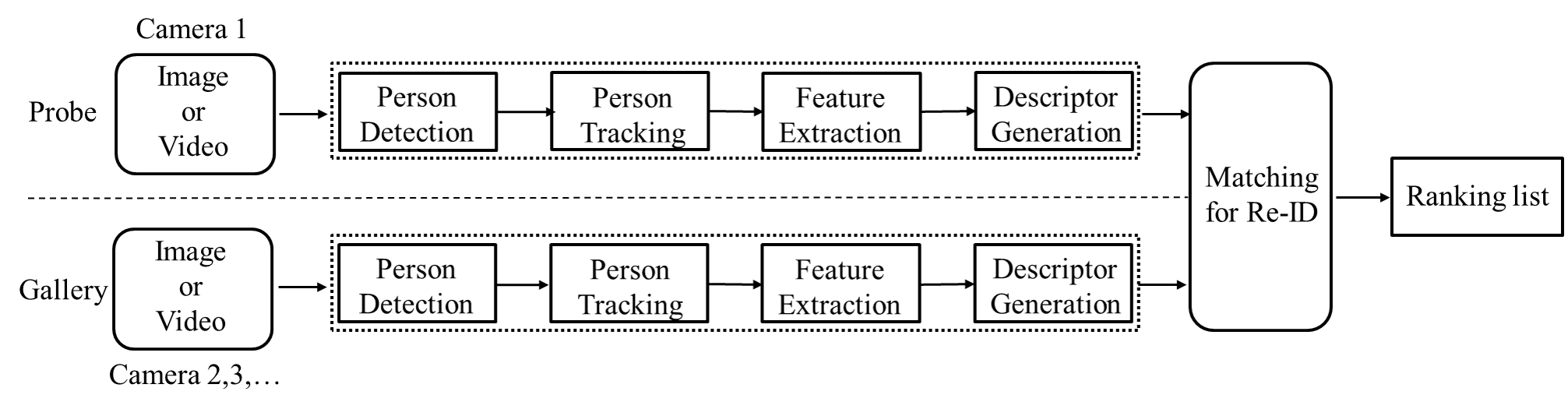}
	\caption{Flowchart of person re-identification system.} 
	\label{fig:1-1-Re-ID-Pipeline}
\end{figure*}

\autoref{fig:1-1-Re-ID-Pipeline} shows the complete flow of the person Re-ID system, which mainly consists of two stages: pedestrian detection and re-identification \citep{Zheng2016Person}. For pedestrian detection, many algorithms with high detection accuracy have emerged, such as YOLO \citep{Redmon2018Yolov3Incremental}, SSD \citep{Liu2016SingleShot} and Fast R-CNN \citep{Girshick2015Fast}. Person Re-ID constructs a large image dataset (Gallery) from the detected pedestrian images and retrieves matching pedestrian images from it using probe images (Probe), so person Re-ID can also be regarded as an image retrieval task \citep{Qi2020RuoJianDu}. The key of person Re-ID is to learn discriminative features of pedestrians to distinguish between pedestrian images with the same identity and those with different identities. However, the difficulty of learning discriminative features of pedestrians is increased by the variation of view, pose, illumination and resolution in different cameras in the real world where pedestrians may appear in multiple cameras in multiple regions.
\begin{figure}[tbh]
	\centering
	\includegraphics[width=0.95\linewidth]{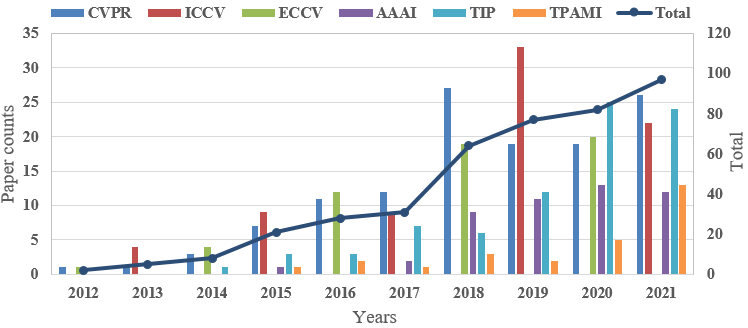}
	\caption{The number of person Re-ID papers on top conferences and journals over the years.} 
	\label{fig:1-1-Re-ID-Papers}
\end{figure}

Traditional person Re-ID methods mainly used manual extraction of fixed discriminative features \citep{Zhong2017Reranking,Zhao2013PersonIdentification,Martinel2015SaliencyWeighted,An2017IntegratingAppearance,Hu2017PersonIdentification} or learned better similarity measures \citep{Dikmen2011PedestrianRecognition,Li2014DeepReID,Chen2015MirrorRepresentation,Wang2016CrossScenario}, which were error-prone and time-consuming, and greatly affected the accuracy and real-time performance of pedestrian Re-ID tasks. In 2014, deep learning was first used in the person Re-ID field \citep{2014DeepYi,2014DeepReIDLi}. \autoref{fig:1-1-Re-ID-Papers} illustrates that there has been a significant increase in the proportion of collected person Re-ID papers over the years. Some researchers designed different loss functions to optimize the learning of discriminative features by network models \citep{Wu2018ExploitUnknown,Zheng2019IdentificationWith,Varior2016GatedSiamese,Hermans2017InDefense,Chen2017BeyondTriplet}. Other researchers extracted more robust features of pedestrians by introducing local feature learning \citep{Sun2018BeyondPart,Sun2019Perceive,Chen2019ABDNet,Miao2019PoseGuided,Zhao2017Spindle} or using attention mechanisms to focus on key information of body parts \citep{Liu2017HydraPlus,Li2018Harmonious,Zhao2017DeeplyLearned,Hu2018SqueezeExcitation,Chen2019MixedHigh,Tay2019AANet}. Ngo et al. \cite{2005MotionNgo,2008BeyondNgo} explored the method of high-level feature extraction aimed to explore context-based concept fusion by modeling inter-concept relationships, which were not modeled based on semantic reasoning. \cite{2012MiningWei}. Several works enhanced the final feature representation by combining global and local features of pedestrians \citep{Wang2018LearningDiscriminative,Wei2017GladGlobal,Su2017PoseDriven,Chen2020SalienceGuided,Zhang2020RelationAware,Zheng2019Pyramidal,Yao2019DeepRepresentation}. Due to the good performance of GAN in generating images and learning features, generative adversarial learning was widely used for person Re-ID tasks \citep{Wei2018PersonTransfer,2020RealHuang,2017Image-ImageDeng,Liu2019AdaptiveTransfer,2020InstanceChen,2018GeneralizingZhong,2018CameraZhong,2017UnlabeledZheng,Yan2019Multi,Zheng2019JointDiscriminative,2018FD-GANGe}. To alleviate the shortage of information in single-frame images, some researchers used the complementary spatial and temporal cues of video sequences to effectively fuse more information in the video sequences \citep{Chung2017StreamSiamese,Liao2019VideoBased,Fu2019SpatialTemporal,Hou2019VrstcOcclusion,Yang2020SpatialTemporal,Yan2020LearningMulti}. Recently, graph convolutional network-based methods \citep{Yang2020SpatialTemporal,Wu2020AdaptiveGraph,Yan2019LearningContext,Shen2018PersonIdentification,Bai2021Unsupervised}  also emerged to learn more discriminative and robust features by modeling graph relationships on pedestrian images. Some researchers \citep{Zheng2020Parameter,Chen2021Learning} improved the robustness of the person Re-ID model by exploiting the information of a person’s 3D shape. These methods are numerous and have different emphases. To give researchers a quick overview of the current state of development and valuable research directions in the field of person Re-ID, we conduct an in-depth survey of deep learning-based person Re-ID methods and summarize the relevant research results in recent years.

\begin{table*}[htbp]
  \centering
  \caption{Comparison of several person Re-ID surveys in recent years.}
    \begin{tabular}{p{17.2em}p{4.19em}p{27.4em}}
    \toprule
    \textbf{Survey} & \textbf{Reference} & \textbf{Contribution} \\
    \midrule
    A survey of approaches and trends in person re-identification\citep{2014BEDAGKARGALASurvey} & IVC’14 & 1. Explored the problem of person Re-ID including system-level challenges, descriptor issues and correspondence issues; 2. Summarized the person Re-ID methods before 2016 including Contextual methods, Non-contextual methods and Active methods. \\
    \midrule
    Person Re-identification  Past, Present and Future\citep{Zheng2016Person} & arXiv’16 & 1. Introduced the history of person re-ID and its relationship between image classification and instance retrieval; 2. Surveyed a broad selection of the hand-crafted systems and the large-scale methods in both image- and video-based re-ID. \\
    \midrule
    A study on deep convolutional neural network based approaches for person re-identification\citep{Chahar2017StudyDeep} & PRMI’17 & 1. Introduced image-based and video-based person Re-ID methods; 2. Discussed some important but undeveloped issues and future research directions. \\
    \midrule
    Survey on person re-identification based on deep learning\citep{Wang2018SurveyPerson} & CAAI’18 & 1. Some deep learning-based person Re-ID methods such as CNN-based, GAN-based and Hybrid-based methods are presented; 2. Proposed the direction of further research. \\
    \midrule
    A systematic evaluation and benchmark for person re-identification: Features, metrics, and datasets\citep{karanam2019SystematicEvaluation} & TPAMI’19 & 1. Presented an extensive review and performance evaluation of single-shot and multi-shot Re-ID algorithms. 2. Introduced the most recent advances in both feature extraction and metric learning. \\
    \midrule
    Beyond intra-modality discrepancy: A comprehensive survey of heterogeneous person re-identification\citep{Wang2020BeyondIntra} & arXiv’19 & 1. Surveyed the models that have been widely employed in heterogeneous person Re-ID. 2. Considered four cross-modality application scenarios: Low-resolution (LR), Infrared (IR), Sketch, and Text. \\
    \midrule
    Deep learning-based methods for person re-identification: A comprehensive review\citep{Wu2019DeepLearning} & NC’19 & 1. Reviewed six types of methods of person Re-ID based on deep learning, including identification deep model, verification deep model, distance metric-based deep model, part-based deep model, video-based deep model and data augmentation-based deep model.\\
    \midrule
    A Brief Survey of Deep Learning Techniques for Person Re-identification\citep{Mathur2020BriefSurvey} & ICETCE’20 & 1. Presented the issues of person Re-ID and the approaches used to solve these issues. \\
    \midrule
    A Survey of Open-World Person Re-Identification\citep{Leng2020SurveyOpen} & TCSVT’20 & 1. Analyzed the discrepancies between closed-word and open-world scenarios. 2. Described the developments of open-set Re-ID works and their limitations. \\
    \midrule
    Person search: New paradigm of person re-identification: A survey and outlook of recent works\citep{Islam2020PersonSearch} & IVC’20 & 1. Discussed about feature representation learning and deep metric learning with novel loss functions. \\
    \midrule
    Survey on Reliable Deep Learning-Based Person Re-Identification Models: Are We There Yet?\citep{Lavi2020SurveyReliable} & arXiv’20 & 1. Surveyed state-of-the-art DNN models being used for person Re-ID task. 2. Discussed their limitations that can work as guidelines for future research. \\
    \midrule
    Person re-identification based on metric learning: a survey\citep{Lavi2020SurveyReliable} & MTA’21 & 1. Summarized the research progress of person Re-ID methods based on metric learning. \\
    \midrule
    Deep Learning for Person Re-identification: A Survey and Outlook\citep{Ye2020Deep} & TPAMI’21 & 1. Reviewed for closed-world person Re-ID from three different perspectives, including deep feature representation learning, deep metric learning and ranking optimization. \\
    \midrule
    Survey on Unsupervised Techniques for Person Re-Identification\citep{Yang2021SurveyUnsupervised} & CDS’21 & 1. Reviewed the state-of-the-art unsupervised techniques of person Re-ID. \\
    \midrule
    SSS-PR: A short survey of surveys in person re-identification\citep{Yaghoubi2021ShortSurvey} & PRL’21 & 1. Proposed a multi-dimensional taxonomy to categorize the most relevant researches according to different perspectives. 2. Filled the gap between the recently published surveys. \\
    \midrule
    Cross-Domain Person Re-identification: A Review\citep{Wang2021CrossDomain} & AIC’21 & 1. Reviewed methods of cross-domain person Re-ID.2. Compared the performance of these methods on public datasets. \\
    \bottomrule
    \end{tabular}%
  \label{tab:TableAllSurvey}%
\end{table*}%

Prior to this survey, some researchers \citep{2014BEDAGKARGALASurvey,Zheng2016Person,Chahar2017StudyDeep,Wang2018SurveyPerson,karanam2019SystematicEvaluation,Wang2020BeyondIntra,Wu2019DeepLearning,Mathur2020BriefSurvey,Leng2020SurveyOpen,Islam2020PersonSearch,Lavi2020SurveyReliable,Lavi2020SurveyReliable,Ye2020Deep,Yang2021SurveyUnsupervised,Yaghoubi2021ShortSurvey,Wang2021CrossDomain} also reviewed the person Re-ID field. In \autoref{tab:TableAllSurvey}, we summarize the major contributions of these reviews. Some of these surveys \citep{Zheng2016Person,Chahar2017StudyDeep} summarized image-based and video-based person Re-ID methods. Other surveys \citep{Wang2018SurveyPerson,Wu2019DeepLearning,Mathur2020BriefSurvey,Ye2020Deep,Lavi2020SurveyReliable,Yang2021SurveyUnsupervised} summarized the deep learning-based person Re-ID methods in different dimensions, which developed rapidly after 2014 and became the main research means. Recently, Wang et al. \citep{Wang2021CrossDomain} outlined methods of cross-domain person Re-ID and compared the performance of these methods on public datasets. Yaghoubi et al. \citep{Yaghoubi2021ShortSurvey} proposed a multi-dimensional taxonomy to categorize the most relevant researches according to different perspectives. Zhou et al. \citep{Domain2021Zhou} provided a review to summarize the developments in domain generalization for computer vision over the past decade. Behera et al. \citep{Behera20202Person} reviewed traditional and deep learning person Re-ID methods in both contextual and non-contextual dimensions. Wu et al. \citep{WU2021Deep} proposed new taxonomies for the two components of feature extraction and metric learning on person Re-ID. Behera et al. \citep{Behera2021Futuristic} conceptualized an overview of interpreting various futuristic cues on the IoT platform for achieving person Re-ID. 

However, there are still some improvements to be made in these surveys, which lack the systematic classification and analysis of deep learning-based person Re-ID methods, also miss many discussions parts for person Re-ID. In this paper, compared to the above review, we focus more on metric learning and representation learning of deep learning methods in person Re-ID tasks and complement the latest research methods of recent years. We present an in-depth and comprehensive review of existing deep learning-based methods and discuss their advantages and limitations. We classify deep learning-based person Re-ID methods in terms of metric and representation learning dimensions, including four categories: deep metric learning, local feature learning, generative adversarial learning and sequence feature learning. 

Deep metric learning focused on designing better loss functions for model training. Common loss functions for person Re-ID included: classification loss, verification loss, contrastive loss, triplet loss and quadruplet loss. Representation learning focused on developing feature construction strategies \citep{Bengio2013Representation,Ye2020Deep}. Therefore, we discussed the common feature learning strategies in recent person Re-ID methods, which were mainly in three categories: 1) Local feature learning, it learned part-level local features to formulate a combined representation for each person image; 2) Generative adversarial learning, it learned the image specific style representation or disentangled representation to achieve image-image style transfer or extract invariant features; 3) Sequence feature learning, it learned video sequence representation using multiple image frames and temporal information. In addition, we subdivided the above four categories based on their methodologies and motivations. This classification has a clear structure, which comprehensively reflects the most common deep metric learning methods and various representation learning methods in Re-ID tasks. Therefore, it is suitable for researchers to explore person Re-ID for practical needs. Furthermore, we attempt to discuss several challenges and research directions. 

Specifically, the main contributions of our work are summarized as follows:
\begin{itemize} 
	\itemsep-0.5em 
	\item We summarize the studies of several recently published person Re-ID surveys and complement the latest research methods to systematically classify deep learning-based person Re-ID methods.
	\item We comprehensively review the research methods of recent deep learning-based person Re-ID. Then, we propose a multi-dimensional taxonomy that classifies these methods into four categories according to metric and representation learning, including deep metric learning, local feature learning, generative adversarial learning, and sequence feature learning.
    \item We subdivide the above four categories based on their methodologies and motivations, discussing the advantages and limitations of part subcategories. This classification is more suitable for researchers to explore these methods from their practical needs.
	\item We summarize some existing challenges in the person Re-ID field and consider that there is still enough necessity to research it. Furthermore, we discuss seven possible research directions for person Re-ID researchers.
\end{itemize} 

The remaining parts of this survey are structured as follows. In \autoref{sec:DatasetsandEvaluationMetrics}, we discuss the common datasets and evaluation metrics used for person Re-ID benchmarks. In \autoref{sec:Methods}, we comprehensively review current deep learning-based methods for person Re-ID and divide them into four categories according to metric learning and representation learning. Furthermore, we subdivide the above four categories based on their methodologies and motivations, discussing the advantages and limitations of part subcategories. Finally, in \autoref{sec:ConclusionAndFutureDirections}, we summarize this paper and discuss the current challenges and future directions in the person Re-ID field.

\section{Datasets and Evaluation metrics}\label{sec:DatasetsandEvaluationMetrics} 
In this section, we present common datasets for evaluating existing deep learning-based person Re-ID methods in both image and video dimensions. In addition, we briefly describe common evaluation metrics for person Re-ID.
\subsection{Datasets}
In recent years, many methods have emerged to improve the performance of person Re-ID. However, the uncertainty of the real world brings about problems including occlusion, lighting changes, camera view switching, pose changes and similar clothing still cannot be well solved. These challenges make many algorithms still not available for real-world applications. Therefore, it is crucial to explore large-scale person Re-ID datasets covering more real scenes. As deep learning-based feature extraction methods gradually replace traditional manual feature extraction methods, deep neural networks require a large amount of training data, which has led to the rapid development of large-scale datasets. The types of datasets and annotation methods vary greatly between datasets. Usually, the datasets used for person Re-ID can be divided into two categories, namely image-based and video-based datasets. The following subsection will introduce two types of commonly used datasets.

\subsubsection{Image-based person Re-ID datasets}
\textbf{VIPeR} \citep{Gray2007EvaluatingAppearance} dataset is the first proposed small person Re-ID dataset. The VIPeR contains two viewpoint cameras, each of which captures only one image. The VIPeR uses manually labeled pedestrians and contains 1,264 images for a total of  632 different pedestrians. Each image is cropped and scaled to a size of 128x48. The VIPeR dataset, which features multiple views, poses, and lighting variations, has been tested by many researchers, but it remains one of the most challenging person Re-ID datasets.

\textbf{CUHK01} \citep{Li2012HumanReidentification} dataset has 971 persons and 3,884 manually cropped images, with each person also having at least two images captured in two disjoint camera views. In the CUHK01 dataset, camera A has more variations of viewpoints and poses, and camera B mainly includes images of the frontal view and the back view.

\textbf{CUHK02} \citep{Li2013Locally} dataset has 1,816 persons and 7,264 manually cropped images. The CUHK02 contains five pairs of camera views(ten camera views)), with each person also having at least two images captured in two disjoint camera views. Compared to the CUHK01 \citep{Li2012HumanReidentification} dataset, the CUHK02 dataset has many identities and camera views and can obtain more configurations(which are the combinations of viewpoints, poses, image resolutions, lightings and photometric settings) of pedestrian images.

\textbf{CUHK03} \citep{Li2014DeepReID} dataset belongs to the large-scale person Re-ID dataset and is collected at the Chinese University of Hong Kong. The CUHK03 is acquired by 10 (5 pairs) cameras and provided with a manual marker and a deformable part model (DPM) detector \citep{Felzenszwalb2010ObjectDetection} together to detect pedestrian bounding boxes. The CUHK03 contains 1,360 different pedestrians with a total of 13,164 images, each of variable size. The CUHK03 improves on the CUHK01 \citep{Li2012HumanReidentification} and CUHK02 \citep{Li2013Locally} by increasing the number of cameras and captured images, thus capturing pedestrian images from more viewpoints. The CUHK03 dataset uses the pedestrian detection algorithm DPM to annotate pedestrians, making it more compatible with person Re-ID in the real world than using individual manual annotations.

\textbf{Market-1501} \citep{Zheng2015Scalable} dataset is a large-scale person Re-ID dataset published in 2015, which was acquired using five high-resolution cameras and one low-resolution camera in front of a supermarket at Tsinghua University. The Market-1501 uses a pedestrian detector DPM to automatically detect pedestrian bounding boxes. It contains 1,501 different pedestrians with a total of 32,668 images, each with a size of 128x64. Compared with CUHK03, Market-1501 has more annotated images and contains 2793+500k interfering factors, and it is closer to the real world.

\textbf{DukeMTMC-reID} \citep{Zheng2017Unlabeled} dataset belongs to a subset of the MTMCT dataset DukeMTMC \citep{Ristani2016Performance}. The DukeMTMC-reID dataset is collected at Duke University using eight static HD cameras. It contains 16,522 training images (from 702 persons), 2,228 query images (from other 702 persons), and a search gallery (Gallery) of 17,661 images.

\textbf{MSMT17} \citep{Wei2018PersonTransfer} dataset is a large-scale person Re-ID dataset published in 2018 and is captured at the campus by fifteen cameras. The MSMT17 uses the pedestrian detector Faster R-CNN \citep{Ren2015FasterTowards} to automatically detect pedestrian-labeled frames. It contains 4,101 different pedestrian information with a total of 126,441 images, which is one of the large datasets of pedestrian and annotated images in the current person Re-ID task. The MSMT17 dataset can cover more scenes than earlier datasets where a single scene and no significant light changes existed.

\textbf{Airport} \citep{karanam2019SystematicEvaluation} dataset is constructed using video data from the six cameras installed at the postcentral security checkpoint at a commercial airport within the United States. The Airport dataset consists of 9,651 identities, 31,238 distractors, a total of 39,902 images, and each one is cropped and scaled to a size of 128x64. The Airport uses pre-detected bounding boxes generated using aggregated channel features(ACF) \citep{Dollar2014Fast} detector, which can accurately reflect real-world person Re-ID issues.

\autoref{tab:imageTable} shows the details of the above datasets. Most of the earlier image-based person Re-ID datasets (VIPeR, CUHK01, CUHK02, CUHK03, Market-1501) have the following limitations: (1) covering a single scene; (2) short time span without significant illumination variations; (3) expensive manual annotation or outdated automatic annotation with DPM detection. The CUHK03 contains viewpoint variations, detection errors, occlusions images. The Market-1501 contains viewpoint variations, detection errors and low-resolution images. But their simulation of the real world is relatively weak. The DukeMTMC-reID dataset contains more challenging attributes include viewpoint variations, illumination variations, detection errors, occlusions, and background clutter. The MSMT17 dataset collects images captured by 15 cameras for both indoor and outdoor scenes.
Therefore, it presents complex scene transformations and backgrounds. The videos cover a long time, thus presenting complex lighting variations. The Airport dataset contains a large number of annotated identities and bounding boxes. To our best knowledge, MSMT17 and Airport are the largest and most challenging public datasets for person Re-ID.

\begin{table*}[t]
	\centering
	\caption{Typical image-based person Re-ID datasets.}
    \begin{tabular}{cp{9.765em}ccccp{10.375em}p{4.625em}}
    \toprule
    \multicolumn{1}{p{5.375em}}{Data type} & Datasets & \multicolumn{1}{p{3em}}{ Years} & \multicolumn{1}{p{2.625em}}{ID} & \multicolumn{1}{p{3.25em}}{Boxes} & \multicolumn{1}{p{3.25em}}{Cameras} & Labeled & Evaluation \\
    \midrule
    \multicolumn{1}{c}{\multirow{6}[2]{*}{Image}} & ViPeR\citep{Gray2007EvaluatingAppearance} & 2007  & 632   & 1,264  & 2     & Handcrafted & CMC \\
          & CUHK01\citep{Li2012HumanReidentification} & 2012  & 971  & 3,884 & 2    & Handcrafted & CMC \\
          & CUHK02\citep{Li2013Locally} & 2013  & 1,816  & 7,264 & 10    & Handcrafted & CMC \\
          & CUHK03\citep{Li2014DeepReID} & 2014  & 1,360  & 13,164 & 10    & DPM+Handcrafted & CMC+mAP \\
          & Market-1501\citep{Zheng2015Scalable} & 2015  & 1,501  & 32,217 & 6     & DPM+Handcrafted & CMC+mAP \\
          & DukeMTMC-reID\citep{2017UnlabeledZheng} & 2017  & 1,812  & 36,441 & 8     & Handcrafted & CMC+mAP \\
          & MSMT17\citep{Wei2018PersonTransfer} & 2018  & 4,101  & 126,441 & 15    & Faster RCNN & CMC+mAP \\
          & Airport\citep{karanam2019SystematicEvaluation} & 2019  & 9,651  & 39,902 & 6     & ACF   & CMC+mAP \\
    \bottomrule
    \end{tabular}%
	\label{tab:imageTable}%
\end{table*}%
\begin{table*}[htbp]
  \centering
  \caption{Typical video-based person Re-ID datasets.}
    \begin{tabular}{cp{9.765em}ccccp{10.375em}p{4.625em}}
    \toprule
    \multicolumn{1}{p{5.375em}}{Data type} & Datasets & \multicolumn{1}{p{3em}}{ Years} & \multicolumn{1}{p{2.625em}}{ID} & \multicolumn{1}{p{3.25em}}{Tracks} & \multicolumn{1}{p{3.25em}}{Cameras} & Labeled & Evaluation \\
    \midrule
    \multicolumn{1}{c}{\multirow{5}[2]{*}{Video}} & PRID-2011\citep{Hirzer2011PersonIdentification} & 2011  & 934   & 400   & 2   & Handcrafted & CMC\\
         & iLIDS-VID \citep{Wang2014PersonIdentification} & 2014  & 300   & 600   & 2     & Handcrafted & CMC \\
          & MARS\citep{Zheng2016MarsVideo} & 2016  & 1,261  & 20,715 & 6   & DPM+GMMCP & CMC+mAP \\
          & DukeMTMC-V\citep{Wu2018ExploitUnknown} & 2018  & 1,812  & 4,832  & 8  & DPM   & CMC+mAP \\
          & LPW\citep{Song2018RegionBased} & 2018  & 2,731  & 7,694  & 11  & DPM+NN+Handcrafted & CMC+mAP \\
    \bottomrule
    \end{tabular}%
  \label{tab:videoTable}%
\end{table*}%
\begin{figure*}[h]
	\centering
	\subfigure[Market-1501\citep{Zheng2015Scalable} ]{
		\includegraphics[width=0.31\linewidth]{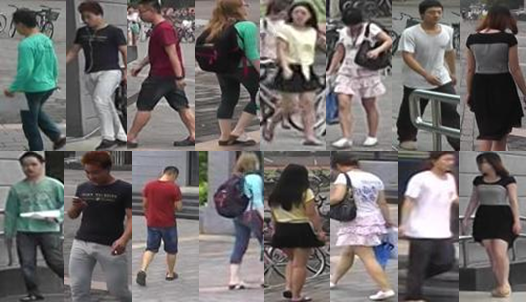}	
	}
	\centering
	\subfigure[DukeMTMC-reID\citep{Zheng2017Unlabeled}]{
		\includegraphics[width=0.31\linewidth]{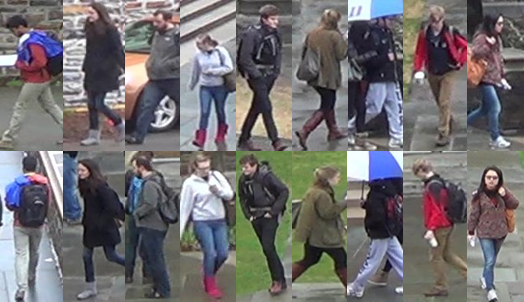}
	}
	\centering
	\subfigure[MSMT17\citep{Zhong2017Reranking}]{
		\includegraphics[width=0.31\linewidth]{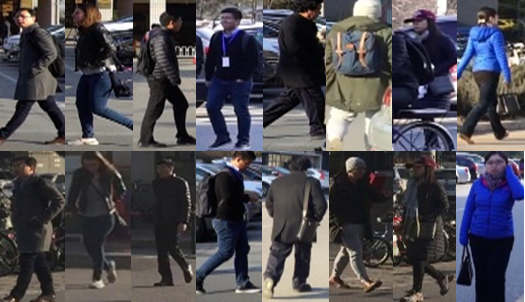}
	}
	\caption{Sampled person images of person Re-ID datasets.}
	\label{fig:2-1-Re-ID-Datasets}
\end{figure*}


\subsubsection{Video-based person Re-ID datasets}
\textbf{PRID2011} \citep{Hirzer2011PersonIdentification} dataset is a video-based person Re-ID dataset proposed in 2011 and is obtained from two non-overlapping camera acquisitions and contains a total of 24,541 images of 934 different pedestrians with manually labeled pedestrian bounding boxes. The resolution size of each image is 128×64.

\textbf{iLIDS-VID} \citep{Wang2014PersonIdentification} dataset is acquired by two cameras at the airport and contains 300 pedestrians with 600 tracks totaling 42,495 pedestrian images. The iLIDS-VID uses manual annotation of pedestrian bounding boxes.

\textbf{MARS} \citep{Zheng2016MarsVideo} dataset is the first large-scale video-based person re-identification dataset proposed in 2016, which contains 1,261 different pedestrians and approximately 20,000 video sequences of pedestrian trajectories acquired from six different cameras. The DPM detector and the generalized maximum multi-cornered problem (GMMCP) tracker \citep{Dehghan2015GmmcpTracker} were used for pedestrian detection and trajectory tracking of MARS, respectively. The MARS dataset contains 3,248 interfering trajectories and is fixed with 631 and 630 different pedestrians to divide the training and test sets, respectively. It can be considered as an extension of Market-1501.

\textbf{DukeMTMC-VideoReID} \citep{Wu2018ExploitUnknown} dataset belongs to a subset of the MTMCT dataset DukeMTMC \citep{Ristani2016Performance} for video-based person Re-ID. This dataset contains 1,812 different pedestrians, 4,832 pedestrian trajectories totaling 815,420 images, in which 408 pedestrians as interference terms, 702 pedestrians for training, 702 pedestrians for testing.

\textbf{LPW(Labeled Person in the Wild)} \citep{Song2018RegionBased} dataset is a largescale video sequence-based person Re-ID dataset that collects three different crowded scenes containing 2,731 different pedestrians and 7,694 pedestrian trajectories with more than 590,000 images. The LPW dataset is collected in crowded scenes and has more occlusions, providing more realistic and challenging benchmarks.

\autoref{tab:videoTable} shows the details of video-based person Re-ID datasets. PRID2011 and iLIDS-VID use only two cameras to capture video and are labeled with fewer identities. That means other identities are only single camera’s frame segments and the lighting and shooting angles of these identities in this dataset may not change much. MARS and DukeMTMC-ViedeReID are large-scale video-based person Re-ID datasets. Their bounding boxes and tracks are automatically generated and contain several natural detection or tracking errors, and each tag may have multiple tracks. LPW is one of the most challenging video-based person Re-ID datasets available and closer to the real world,  distinguishes from existing datasets in three aspects: more identities and tracks, automatically detected bounding boxes, and far more crowded scenes with a larger time span. 

\autoref{fig:2-1-Re-ID-Datasets} shows some sample images of a partial person Re-ID dataset. We can see that with the development of large-scale person Re-ID datasets, the number of pedestrian IDs and the number of labeled frames or trajectories in the datasets are increasing, and the scenarios covered by the datasets are getting richer. These datasets use a combination of deep learning detectors and manual annotation to detect pedestrian bounding boxes, making the latest datasets closer and closer to the real world, thus enhancing the robustness of person Re-ID models. In addition, almost all mainstream person Re-ID datasets are evaluated using mean average precision (mAP) and cumulative matching characteristics (CMC) curves for performance evaluation.
\subsection{Evaluation metrics} 
The commonly used evaluation metrics of person Re-ID algorithms are cumulative matching characteristics (CMC) curves and mean average precision (mAP).

In pattern recognition systems, CMC curves are important evaluation metrics in the fields of face, fingerprint, iris detection and person Re-ID, which can comprehensively assess the merits of model algorithms. Furthermore, CMC curves are considered to be a comprehensive reflection of the performance of the person Re-ID classifier. Before calculating the CMC curves, the probability ${Acc_k}$ that the top-k retrieved images (top-k) in the gallery contain the correct query result is obtained by ranking the similarity between the query target and the target image to be queried, which is calculated as follows: 
\begin{equation}\label{eqn:Lyapdisturb_ACCK}
{Acc_k}=\begin{array}{l} 
\begin{cases}
  1& \text{if top-k rank gallery samples}\\
   & \text{ contain query identity.} \\
  0& \text{otherwise} 
\end{cases} 
\end{array} 
\end{equation}
CMC curves are calculated by adding up the ${Acc_k}$ of each query image and dividing it by the total number of query images, which are usually expressed as Rank-k. For example, Rank-1 accuracy indicates the probability of correctly matching to the first target in the matching list.

A single evaluation metric often cannot comprehensively evaluate the comprehensive performance of the person Re-ID algorithm. The mAP can reflect the extent to which all images with correct queries are at the front of the result queue in the query results. Considering  both the average precision (AP) and precision-recall curve (PR) of the query process \citep{Zheng2015Scalable}, instead of just focusing on the hit rate, which can measure the performance of person Re-ID algorithms more comprehensively. The algorithm is usually necessary to be evaluated separately for CMC curves and mAP in person Re-ID tasks. Zheng et.al. \citep{Zhong2017Reranking} proposed the Re-ranking method, which can re-rank the query results and further improve the effectiveness of Rank-k and mAP accuracy.
\section{Deep Learning Based Re-ID Method}\label{sec:Methods} 
\begin{figure*}[tbh]
	\centering
	\includegraphics[width=0.95\linewidth,height=0.5\linewidth]{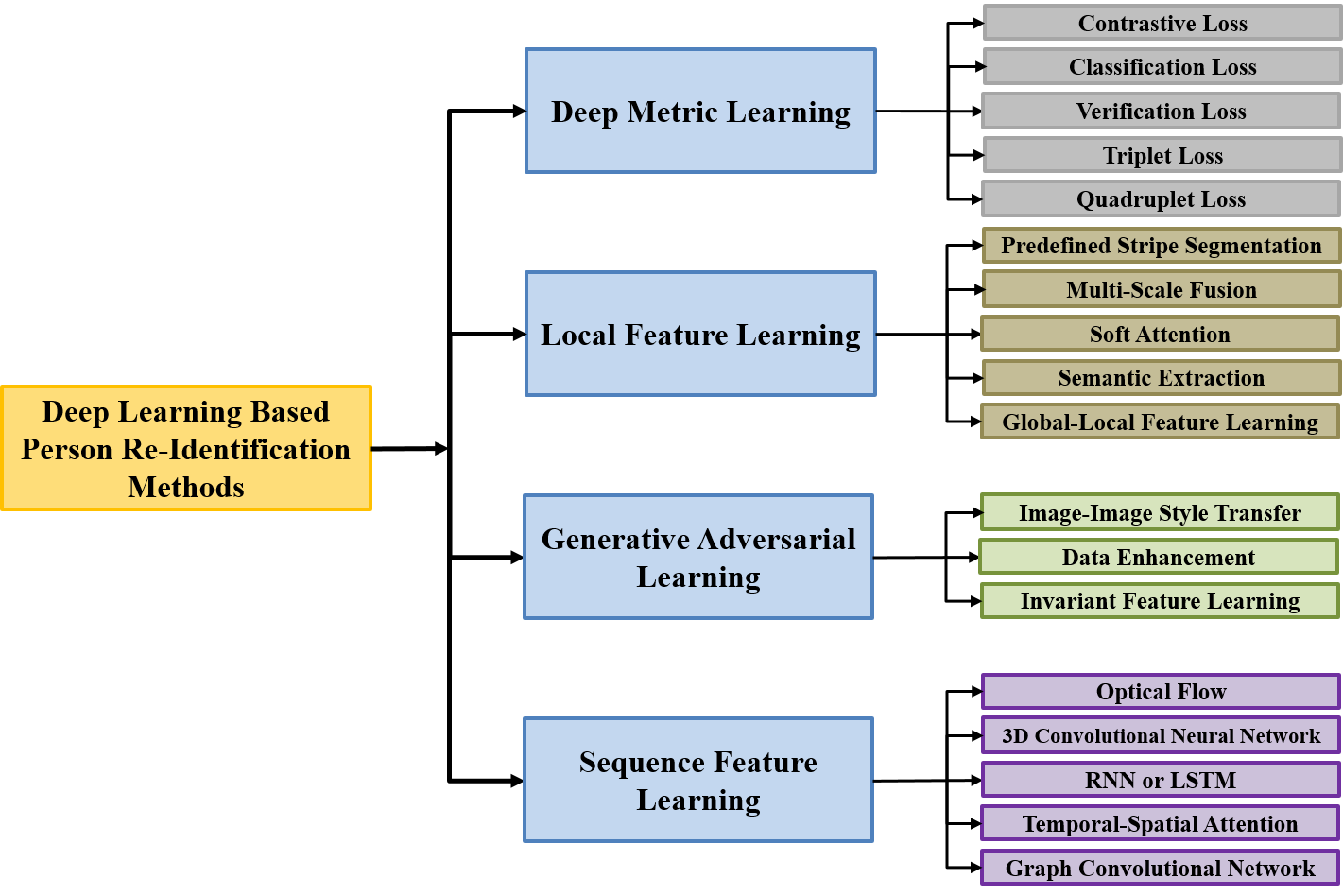}
	\caption{Classification structure of deep learning-based person re-identification methods.}
	\label{fig:3-0-Structure}
\end{figure*}

In this section, we classify deep learning-based person Re-ID methods into four categories with classification structure is shown in \autoref{fig:3-0-Structure}, including methods for depth metric learning, local feature learning, generative adversarial learning and sequences feature learning. In addition, we subdivide the above four categories according to their methodologies and motivations, discussing and comparing the advantages and limitations of part subcategories.

\subsection{Deep metric learning}
\begin{figure*}[tbh]
	\centering
	\includegraphics[width=0.95\linewidth]{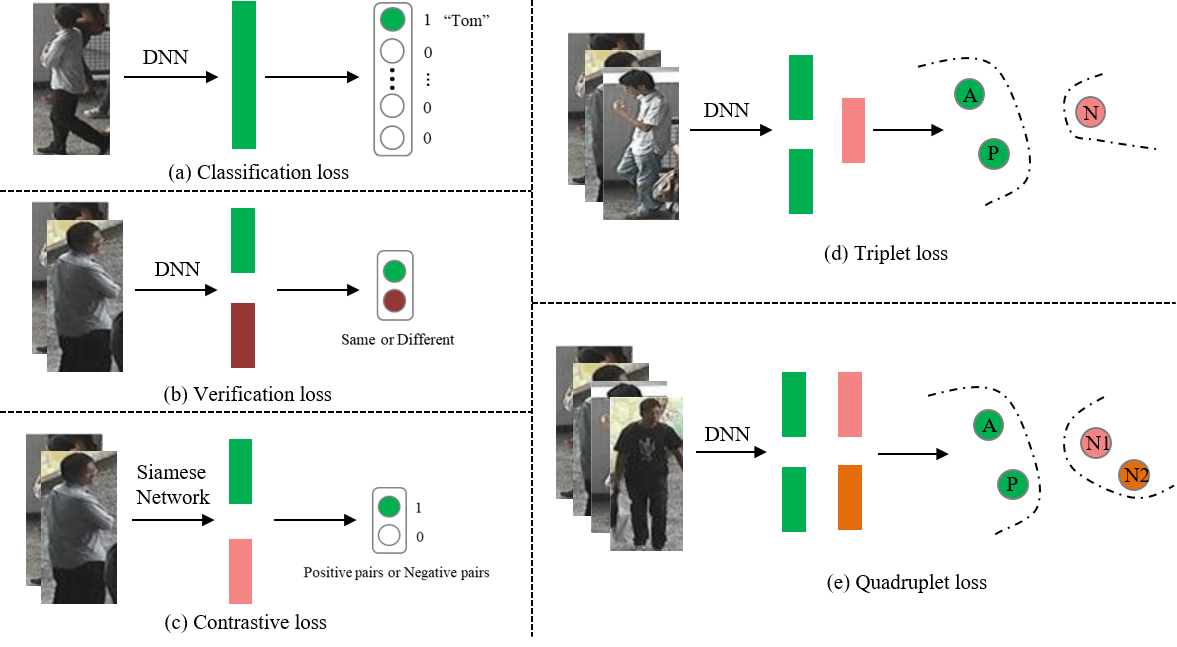}
	\caption{An illustration shows a variety of existing deep metric learning-based person Re-ID losses. (a) Classification loss. (b) Verification loss. (c) Contrastive loss. (d) Triplet loss,  where $A$, $P$, $N$ indicate anchor, positive and negative samples, respectively. (e) Quadruplet loss, where $N1$ and $N2$ are different negative samples.} 
	\label{fig:3-1-Loss}
\end{figure*}

Deep metric learning (DML) is one of the metric learning (ML) methods that aims to learn the similarity or dissimilarity between two pedestrian objects. The main goal of DML is to learn a mapping from the original image to the feature embedding (FE) such that the same pedestrians have smaller distances using a distance function on the feature space and different pedestrians feature farther apart from each other \citep{Song2016DeepMetricLearning,Duan2018DeepLocalized}. With the rise of deep neural networks(DNNs), DML has been widely used in computational vision, such as face recognition, image retrieval, and person Re-ID. DML is mainly used to constrain the learning of discriminative features by designing loss functions for network models \citep{Ye2020Deep}. In this paper, we focus on loss functions commonly used in person Re-ID tasks, including classification loss \citep{Wu2018ExploitUnknown,Zheng2019IdentificationWith,2017Image-ImageDeng,2018CameraZhong,Huang2018AdversariallyOccluded,Zhong2019InvarianceMatters,Luo2019SpectralFeature}, verification loss \citep{Zheng2019IdentificationWith,Zheng2017Discriminatively,Chen2018DeepTransfer,Ye2020Deep}, contrastive loss \citep{Varior2016GatedSiamese,Varior2016Siamese,Wang2018PersonIdentification}, triplet loss \citep{Hermans2017InDefense,Ristani2018Multitarget,Song2018MaskGuided} and quadruplet loss \citep{Chen2017BeyondTriplet}.  An illustration
of five loss functions is shown in \autoref{fig:3-1-Loss}. These deep metric learning methods enable models to learn discriminative features automatically, which can solve the problem of manually designing features that consume a lot of labor costs.

\subsubsection{Classification loss}
Zheng et al. \citep{Zheng2016Person,Zheng2017Wild} treated the training process of person Re-ID as a multi-classification task for images and proposed an ID-discriminative embedding (IDE) network. The IDE treats each pedestrian as a different class and uses the ID of the pedestrian as a classification label to train a deep neural network, so the classification loss is also called ID loss. The training network for classification loss usually inputs a picture and connects a fully connected layer (FC) for classification at the end of the network, and then maps the feature vectors of the image onto the probability space by the softmax activation function. The cross-entropy loss for multi-classification of the person Re-ID task can be expressed as: 
\begin{equation}\label{eqn:Lyapdisturb_ID}
\begin{aligned}
\mathcal{L}_{id}=-\sum_{a=1}^{K} q\left ( x_a \right ) \log_{}{} p\left ( y_a|x_a \right )
\end{aligned}
\end{equation}

Where $K$ represents the number of training sample ID categories per batch, $q(x_a)$ denotes the label of sample image $x_a$. If $x_a$ is identified as $y_a$, then $q(x_a)=1$, otherwise $q(x_a)=0$. $p(y_a|x_a)$ is the probability that picture $x_a$ is predicted as category $y_a$ using the softmax activation function. Classification loss is widely used as a depth metric learning for person Re-ID methods because of its advantages such as easy training of models and mining hard samples \citep{Wu2018ExploitUnknown,Zheng2019IdentificationWith,2017Image-ImageDeng,2018CameraZhong,Huang2018AdversariallyOccluded,Zhong2019InvarianceMatters,Luo2019SpectralFeature}. However, using ID information alone is not enough to learn a model with sufficient generalization ability. Therefore, ID loss usually needs to be combined with other losses to constrain the training of the model.

\subsubsection{Verification loss}
Person Re-ID can also be treated as a validation problem, and validation loss is proposed to guide the training of the model. In contrast to classification loss, the network trained by verification loss requires two images as input, and a binary loss is computed by fusing the feature information of the two images, which in turn determines whether the input two images are the same pedestrian \citep{Zheng2019IdentificationWith,Zheng2017Discriminatively,Chen2018DeepTransfer,Ye2020Deep}. The expression of the cross-entropy validation loss function is as follows: 
\begin{equation}\label{eqn:Lyapdisturb_Ver}
\begin{aligned}
\mathcal{L}_{v}=-y_{ab} \log_{}{} p\left ( y_{ab}|f_{ab}\right )-\\(1-y_{ab}) \log_{}{} \left (1-p(y_{ab}|f_{ab})\right ) 
\end{aligned}
\end{equation}

Supposing the network inputs two images $x_a$ and $x_b$, we get the feature vectors $f_a$ and $f_b$ of these two images respectively, and calculate the difference feature $f_{ab}={(f_a-f_b)^{2}}$ of the two feature vectors. We use the softmax activation function to calculate the probability $p$ that the image pairs $x_a$ and $x_b$ have the same pedestrian ID, where $y_{ab}$ is the pedestrian ID label of the two images. When the images $x_a$ and $x_b$ have the same ID, $y_{ab}=1$, otherwise, $y_{ab}=0$.

The verification loss is less efficient in recognition because it can only input a pair of images to judge the similarity when tested while ignoring the relationship between the image pair and other images in the dataset. For this reason, researchers considered combining classification and validation networks \citep{Zheng2017Discriminatively,Chen2018DeepTransfer}, and the combined loss can be expressed as $\mathcal{L}=\mathcal{L}_{id}+\mathcal{L}_{v}$. The hybrid loss can combine the advantages of classification loss and verification loss, which can predict the identity ID of pedestrians and perform similarity metrics simultaneously, thus improving the accuracy of person Re-ID.

\subsubsection{Contrastive loss}
Contrastive loss, which mainly constrains the similarity or dissimilarity between pairs of data, is generally used for model training of twin networks (Siamese Network) in person Re-ID tasks \citep{Varior2016GatedSiamese,Varior2016Siamese,Wang2018PersonIdentification}. Its function can be expressed as:
\begin{equation}\label{eqn:Lyapdisturb_Con}
\begin{aligned}
\mathcal{L}_{c}=yd\left( x_a-x_b \right)^2+\left ( 1-y \right )  \left [m-d\left( x_a-x_b \right)^2  \right ] _+
\end{aligned}
\end{equation}

Where $\left [ z \right ]_+ =max(0,z)$, $x_a$ and $x_b$ are two images input to the twin network at the same time. $d(x_a,x_b)$ usually indicates the euclidean distance (similarity) of the two images. $m$ is the set training threshold, and $y$ is the label of whether each pair of training images matches. When $y=1$, it means that the input images $x_a$ and $x_b$ belong to the pedestrians with the same ID (positive sample pair). When $y=0$, it means that the input images $x_a$ and $x_b$ belong to pedestrians with different IDs (negative sample pair). $\mathcal{L}_{c}$ reflects well the matching degree of sample pairs, which is often used to train models for person Re-ID feature extraction and often works together with classification loss combinations for training networks \citep{2017Image-ImageDeng}.

\subsubsection{Triplet loss}
Triplet loss is one of the most widely used depth metric losses in person Re-ID tasks, and it aims to minimize the intra-class distance and maximize the inter-inter-class distance of samples. With the development of deep neural networks, a large number of variants based on triplet loss have emerged \citep{Hermans2017InDefense,Ristani2018Multitarget,Song2018MaskGuided,Li2021Combined}. The triplet loss function can be expressed as:
\begin{equation}\label{eqn:Lyapdisturb_Trip}
\begin{aligned}
\mathcal{L}_{trip}=\left [ m+d(x_a,x_p)-d(x_a,x_n) \right ]_+
\end{aligned}
\end{equation}

Different from contrast loss, the input of triplet loss is a triplet consisting of three images. Each triplet contains a pair of positive samples and a negative sample, where $x_a$ is the Anchor image, $x_p$ is the Positive image, and $x_n$ is the Negative image, and the pedestrians of $x_a$ and $x_p$ have the same ID. The pedestrians of $x_a$ and $x_n$ have different IDs. By model training, the distance between $x_a$ and $x_p$ in the Euclidean space is made closer than the distance between $x_n$ and $x_a$. To improve the performance of models, some deep learning-based person Re-ID methods use a combination of classification loss and triplet loss \citep{Guo2019BeyondHuman,Chen2019SelfCritical,Liu2019DeepReinforcement,Song2019GeneralizablePerson,Zhou2019DiscriminativeFeature}. Experiments have shown that combining these two losses facilitates the model to learn discriminative features. Traditional triplet loss randomly selects three images from the training set during training, which may result in a simple combination of samples and lacks the training of hard sample combinations and makes the training model less generalizable. For this reason, some researchers considered improving triplet loss for mining hard samples \citep{Hermans2017InDefense,Shi2016Embedding,Mishchuk217WorkingHard}.

\subsubsection{Quadruplet loss}
Another improvement for the triplet loss is to add a negative sample picture $X_{n2}$ to form a quadruplet loss \citep{Chen2017BeyondTriplet}, where negative sample $X_{n1}$ and negative sample $X_{n2}$ have different pedestrian IDs. The expression of the quadruplet loss function is:
\begin{equation}\label{eqn:Lyapdisturb_quad}
\begin{aligned}
\mathcal{L}_{quad}=\left [ m_1+d(x_a,x_p)-d(x_a,x_{n1}) \right ]_++\\
\left [ m_2+d(x_a,x_p)-d(x_{n1},x_{n2}) \right ]_+
\end{aligned}
\end{equation}

Where $m_1$ and $m_2$ are custom training thresholds. The positive and negative sample pairs have the same anchored image $x_a$. The first term of $\mathcal{L}_{quad}$ is identical to the triplet loss function, which is used to constrain the relative distance between positive and negative sample pairs. The traditional triplet loss function often increases the inter-class distance of negatives sample pairs, which affects the feature learning of image $x_a$. For this reason, $\mathcal{L}_{quad}$ introduces a second term to constrain the absolute distance between positive and negative sample pairs. The positive and negative sample pairs in the second term have different anchor images, which can effectively reduce the intra-class distance of positive sample pairs while increasing the distance of negative sample pairs between classes. In order to make the  first term play a dominant role, it is usually important to ensure that ${m}_{1}>{m}_{2}$ during the training process. However, most person Re-ID methods using triplet loss drive focused more on differentiating appearance differences and cannot effectively learn fine-grained features. To address this issue, Yan et al. \citep{Yan2021Beyond} introduce a novel pairwise loss function that enables Re-ID models to learn the fine-grained features by adaptively enforcing an exponential penalization. 
\subsection{Local feature learning}
Based on the features extracted from pedestrian images for classification, person Re-ID methods can be classified into global feature learning-based methods and local feature learning-based methods. The global feature learning methods usually extract one feature of the pedestrian image \citep{Wu2016PersonnetPerson,Wang2016JointLearning,Qian2017MultiScale}, and it is difficult for this method to capture the detailed information of the pedestrian. Therefore, how to extract discriminative local features of pedestrians with subtle differences becomes a problem for researchers to focus on.

The local feature learning-based methods aim at learning pedestrian discriminative features and ensuring the alignment of each local feature. Manual annotation or neural networks are usually used to automatically focus on certain local regions with key information and extract the distinguishing features from these regions. Commonly used local feature learning methods are predefined stripe segmentation \citep{Sun2018BeyondPart,Sun2019Perceive,Miao2019PoseGuided,Varior2016Siamese,Liu2018CA3NetLiu,Fu2019Horizontal}, multi-scale fusion \citep{Liu2016MultiScale,Chen2017PersonIdentification,Zhou2019OmniScale,Li2017LearningDeep,Yang2019TowardsRich}, soft attention \citep{Chen2019ABDNet,Liu2017HydraPlus,Li2018Harmonious,Zhao2017DeeplyLearned,Hu2018SqueezeExcitation,Chen2019MixedHigh,Tay2019AANet,Zhang2021Heterogeneous,Ning2021JWSAA}, pedestrian semantic extraction \citep{Guo2019BeyondHuman,Miao2019PoseGuided,Zhao2017Spindle,Zhao2017DeeplyLearned,Kalayeh2018HumanSemantic} and global-local feature learning \citep{Wang2018LearningDiscriminative,Wei2017GladGlobal,Su2017PoseDriven,Chen2020SalienceGuided,Zhang2020RelationAware,Zheng2019Pyramidal,Yao2019DeepRepresentation}. These methods can alleviate the problems of occlusion, boundary detection errors, view, and pose variations. 

\subsubsection{Predefined stripe segmentation}
The main idea of the method based on predefined stripe segmentation is to strip the learned features according to some predefined division rules, which must ensure that the partitions are spatially aligned. Liu et al. \citep{Liu2018CA3NetLiu} proposed an attribute and appearance-based contextual attention network, where the appearance network learns spatial features from the whole body, horizontal and vertical parts of the pedestrian. Varior et al. \citep{Varior2016Siamese} divided the pedestrian image into several strips uniformly and extracted local features from each strip image block. Sun et al. \citep{Sun2018BeyondPart} considered the content consistency within each stripe to propose a local convolutional baseline (PCB). The PCB uses a uniform feature partitioning strategy to learn local features and outputs convolutional features consisting of multiple stripes to enhance the consistency of each partition's feature content, thus ensuring stripes are spatially aligned.

Although the above methods can extract discriminative features for the striped areas, it may lead to incorrect retrieval results as the model is unable to distinguish between the obscured and unobscured areas. To relieve occlusion, Sun et al. \citep{Sun2019Perceive} proposed a visibility-aware local model based on PCB to ensure that local features are spatially aligned and avoid interference due to pedestrian occlusion by learning common region features that are visible in both images. Fu et al. \citep{Fu2019Horizontal} horizontally slice the deep feature maps into multiple spatial strips using various pyramid scales and used global average pooling and maximum pooling to obtain discriminative features for each strip, which was named horizontal pyramid pooling (HPP). HPP can ignore that interference information, mainly coming from similar clothing or background.

\subsubsection{Multi-scale fusion}
Small-scale feature maps have a strong ability to represent spatial geometric information and can obtain detailed information of the image. Large-scale feature maps are good at characterizing semantic information and can get contour information of images. Extracting pedestrian features at multiple scales for fusion can obtain rich pedestrian feature representation.

Liu et al. \citep{Liu2016MultiScale} proposed a multi-scale triple convolutional neural network, which can capture pedestrian appearance features at different scales. Since pedestrian features learned at different scales differ or conflict, the direct merging of features at multiple scales may not achieve the best fusion effect. Therefore, researchers have started to focus on the complementary advantages of cross-scale implicit associations. Chen et al. \citep{Chen2017PersonIdentification} studied the problem of person Re-ID multi-scale feature learning and proposed a deep pyramidal feature learning deep neural network framework, which can overcome the differences in cross-scale feature learning while learning multi-scale complementary features. Zhou et al. \citep{Zhou2021Learning} presented a Re-ID CNN termed Omni-scale network (OSNet) to learn features that not only captured different spatial scales but also encapsulated a synergistic combination of multiple scales. In traditional person Re-ID datasets, OSNet achieved state-of-the-art performance, despite being much smaller than existing Re-ID models. The challenge of large intra-class variation and small inter-class variation often arises in cross-camera person Re-ID tasks. For example,  cross-camera viewpoint changes can obscure parts of the person with discriminative features, or pedestrians wearing similar clothes appear across cameras, which makes the matching of the same person incorrect.

\subsubsection{Soft attention}
The goal of attention is to find the areas that have a greater impact on the feature map and to focus the model on discriminative local parts of body appearances to correct misalignment, eliminate background perturbance. Due to the good performance of the attention mechanism in the computer vision field, it is often used as a local feature learning in the person Re-ID tasks. Most of the current attention-based person Re-ID methods tend to use soft attention which can be divided into spatial attention, channel attention, mixed attention, non-local attention and position attention. Liu et al. \citep{Liu2017HydraPlus} proposed an attention-based deep neural network capable of capturing multiple attention features from the underlying to the semantic layer to learn fine-grained integrated features of pedestrians. Li et al. \citep{Li2018Harmonious} used balanced attention convolutional neural networks to maximize the complementary information of attention features at different scales to solve the person Re-ID challenge for arbitrary unaligned images. To obtain the local fine-grained features of a person, Ning et al. \citep{Ning2021Feature} proposed a multi-branch attention network with diversity loss, and the local features were obtained via adaptive filtering by removing interference information.

The above spatial attention-based methods tend to focus only on the local discriminative features of pedestrians but ignore the impact of feature diversity on pedestrian retrieval. Chen et al. \citep{Chen2019ABDNet} proposed an attentional diversity network that used complementary channel attention module (CAM) and position attention module (PAM) to learn the characteristics of pedestrian diversity. Considering that features extracted from first-order attention such as spatial attention and channel attention are not discriminative in complex camera view and pose change scenarios \citep{Hu2018SqueezeExcitation}. Chen et al. \citep{Chen2019MixedHigh} proposed a higher-order attention module, which modelled the complex higher-order information in the attention mechanism to mine discriminative attention features among pedestrians.

\subsubsection{Semantic extraction}
Some researchers used deep neural networks to extract semantic information such as body parts or body postures instead of bounding boxes to extract local features from pedestrian body parts to improve the performance of person Re-ID. Zhao et al. \citep{Zhao2017Spindle} considered the application of body structure information to the person Re-ID task and proposed a novel CNN, called SpindleNet. Specifically, firstly, SpindleNet used a body part generation network to locate 14 key points of body parts to extract 7 body regions of the pedestrian. Secondly, SpindleNet captured semantic features from different body regions using a convolutional neural network. Finally, SpindleNet used a tree fusion network with competing strategies to merge the semantic features from different body regions. SpindleNet can align the features of body parts over the whole image and can better highlight local detail information.

Sometimes, not only do the body parts contain discriminative features but also non-body parts may contain certain key features, such as the pedestrian's distinguishing backpack or handbag. Therefore, some researchers considered the alignment of non-body parts. Guo et al. \citep{Guo2019BeyondHuman} proposed a dual part-aligned representation scheme that captured distinguishing information beyond body parts using an attention mechanism to update the representation by exploiting the complementary information from both the accurate human parts and the coarse non-human parts. Miao et al. \citep{Miao2019PoseGuided} proposed a pose-guided feature alignment scheme to distinguish information from occlusion noise by pedestrian pose bounding markers, thus aligning the query image and the non-occluded area of the queried image.

\subsubsection{Global-local feature learning}
Local feature learning can capture detailed information about a region of the pedestrian, but the reliability of local features can be affected by variations in pose and occlusion. Therefore, some researchers often combine fine-grained local features with coarse-grained global features to enhance the final feature representation.
Wang et al. \citep{Wang2018LearningDiscriminative} proposed a multi-granularity feature learning strategy with global and local information, including one branch for global feature learning and two branches for local feature learning. Ming et al. \citep{2021Global-LocalMing} designed a global-local dynamic feature alignment network (GLDFA-Net) framework, which contained both global and local branches. The local sliding alignment(LSA) strategy was introduced into the local branch of GLDFA-Net to guide the computation of distance metrics, which can further improve the accuracy of the testing phase. To mitigate the impact of imprecise bounding boxes on pedestrian matching, Zheng et al. \citep{Zheng2019Pyramidal} proposed a coarse-grained to fine-grained pyramid model that integrates not only local and global information of pedestrians but also progressive cues from coarse to fine-grained. The model can match pedestrian images of different scales and retrieve pedestrian images with the same local identity even when the images are not aligned.

\subsubsection{Comparison and discussion}
\begin{table*}[htbp]
  \centering
  \caption{Comparison of experimental results of local feature learning based methods. $*$ represents the use of multiple deep learning methods. The bold and underlined numbers represent the top two results.}
    \begin{tabular}{cp{6.55em}llllp{2.725em}p{2.725em}p{5.315em}}
    \toprule
    \multicolumn{1}{c}{\multirow{2}[2]{*}{Categories}} & \multirow{2}[2]{*}{Methods} & \multicolumn{2}{p{4.63em}}{CUHK03} & \multicolumn{2}{p{5.94em}}{Market-1501} & \multicolumn{2}{p{6.44em}}{DukeMTMC-reID} & \multirow{2}[2]{*}{Reference} \\
          & \multicolumn{1}{c}{} & \multicolumn{1}{p{2.315em}}{mAP } & \multicolumn{1}{p{2.315em}}{R-1} & \multicolumn{1}{p{3.19em}}{mAP } & \multicolumn{1}{p{2.75em}}{R-1} & mAP   & R-1   & \multicolumn{1}{c}{} \\
    \midrule
    \multicolumn{1}{c}{\multirow{5}[2]{*}{Predefined stripe segmentation}} & PCB+RPP\citep{Sun2018BeyondPart} & \multicolumn{1}{p{2.315em}}{-} & \multicolumn{1}{p{2.315em}}{-} & 81.6  & 93.8  & \multicolumn{1}{l}{69.2} & \multicolumn{1}{l}{83.3} & ECCV'18 \\
          & CA3Net\citep{Liu2018CA3NetLiu} & \multicolumn{1}{p{2.315em}}{-} & \multicolumn{1}{p{2.315em}}{-} & 80.0    & 93.2  & \multicolumn{1}{l}{70.2} & \multicolumn{1}{l}{84.6} & ACMMM'18 \\
          & PGFA\citep{Miao2019PoseGuided}* & \multicolumn{1}{p{2.315em}}{-} & \multicolumn{1}{p{2.315em}}{-} & 76.8  & 91.2  & \multicolumn{1}{l}{65.5} & \multicolumn{1}{l}{82.6} & ICCV'19 \\
          & VPM\citep{Sun2019Perceive} & \multicolumn{1}{p{2.315em}}{-} & \multicolumn{1}{p{2.315em}}{-} & 80.8  & 93.0    & \multicolumn{1}{l}{72.6} & \multicolumn{1}{l}{83.6} & CVPR'19 \\
          & HPM\citep{Fu2019Horizontal} & 57.5  & 63.9  & 82.7  & 94.2  & \multicolumn{1}{l}{74.3} & \multicolumn{1}{l}{86.6} & AAAI'19 \\
    \midrule
    \multicolumn{1}{c}{\multirow{4}[2]{*}{Multi-scale fusion}} & DPFL\citep{Chen2017PersonIdentification} & 40.5  & 43.0    & 72.6  & 88.6  & \multicolumn{1}{l}{60.6} & \multicolumn{1}{l}{79.2} & ICCV'17 \\
          & MSCAN\citep{Li2017LearningDeep} & \multicolumn{1}{p{2.315em}}{-} & 74.2 & 66.7  & 86.8 & -     & -     & CVPR'17 \\
          & OSNet\citep{Zhou2019OmniScale} & 67.8  & 72.3  & 84.9  & 94.8  & \multicolumn{1}{l}{73.5} & \multicolumn{1}{l}{88.6} & ICCV'19 \\
          & CAM\citep{Yang2019TowardsRich} & 64.2  & 66.6  & 84.5  & 94.7  & \multicolumn{1}{l}{72.9} & \multicolumn{1}{l}{85.8} & CVPR'19 \\
    \midrule
    \multicolumn{1}{c}{\multirow{5}[2]{*}{Soft attention}} & HydraPlus\citep{Liu2017HydraPlus} & \multicolumn{1}{p{2.315em}}{-} & \underline{91.8}  & \multicolumn{1}{p{3.19em}}{-} & 76.9  & -     & -     & ICCV'17 \\
          & HA-CNN\citep{Li2018Harmonious} & 44.4  & 41.0    & 75.7  & 91.2  & \multicolumn{1}{l}{63.8} & \multicolumn{1}{l}{80.5} & CVPR'18 \\
          & ABD-net\citep{Chen2019ABDNet} & \multicolumn{1}{p{2.315em}}{-} & \multicolumn{1}{p{2.315em}}{-} & 88.3 & 95.6  & \multicolumn{1}{l}{78.6} & \multicolumn{1}{l}{89.0} & ICCV'19 \\
          & HOA\citep{Chen2019MixedHigh} & 72.4  & 77.2  & 85.0    & 95.1  & \multicolumn{1}{l}{77.2} & \multicolumn{1}{l}{89.1} & ICCV'19 \\
          & AANet\citep{Tay2019AANet} & \multicolumn{1}{p{2.315em}}{-} & \multicolumn{1}{p{2.315em}}{-} & 83.4 & 93.9 & \multicolumn{1}{l}{74.3} & \multicolumn{1}{l}{87.7} & CVPR'19 \\
          
          & HLGAT\citep{Zhang2021Heterogeneous} & \multicolumn{1}{p{2.315em}}{\underline{80.6}} & \multicolumn{1}{p{2.315em}}{83.5} & \textbf{93.4} & \textbf{97.5} & \multicolumn{1}{l}{\textbf{87.3}} & \multicolumn{1}{l}{\textbf{92.7}} & CVPR'21 \\
          & PAT\citep{Li2021Diverse} & \multicolumn{1}{p{2.315em}}{-} & \multicolumn{1}{p{2.315em}}{-} & 88.0 & 95.4 & \multicolumn{1}{l}{78.2} & \multicolumn{1}{l}{88.8} & CVPR'21 \\
          
    \midrule
    \multicolumn{1}{c}{\multirow{3}[2]{*}{Semantic extraction}} & Spindle\citep{Zhao2017Spindle} & \multicolumn{1}{p{2.315em}}{-} & 88.5  & \multicolumn{1}{p{3.19em}}{-} & 76.9  & -     & -     & CVPR'17 \\
          & SPReID\citep{Kalayeh2018HumanSemantic} & \multicolumn{1}{p{2.315em}}{-} & \textbf{94.3} & 83.4 & 93.7 & \multicolumn{1}{l}{73.3} & \multicolumn{1}{l}{85.9} & CVPR'18 \\
          & P2-Net\citep{Guo2019BeyondHuman} & 73.6  & 78.3  & 85.6  & 95.2  & \multicolumn{1}{l}{73.1} & \multicolumn{1}{l}{86.5} & ICCV'19 \\
    \midrule
    \multicolumn{1}{c}{\multirow{6}[1]{*}{Global-Local feature learning}} & GLAD\citep{Wei2017GladGlobal} & \multicolumn{1}{p{2.315em}}{-} & 85.0    & 73.9  & 89.9  & -     & -     & ACMMM'17 \\
          & PDC\citep{Su2017PoseDriven} & \multicolumn{1}{p{2.315em}}{-} & 88.7  & 63.4 & 84.1 & -     & -     & ICCV'17 \\
          & Pyramid\citep{Zheng2019Pyramidal} & 76.9  & 78.9  & 88.2  & 95.7  & \multicolumn{1}{l}{\underline{79.0}} & \multicolumn{1}{l}{89.0} & CVPR'19 \\
          & RGA\citep{Zhang2020RelationAware} & 77.4  & 81.1  & 88.4  & \underline{96.1} & -     & -     & CVPR'20 \\
          & SCSN\citep{Chen2020SalienceGuided} & \textbf{84.0} & 86.8  & \underline{88.5} & 95.7  & \multicolumn{1}{l}{\underline{79.0}} & \multicolumn{1}{l}{\underline{91.0}} & CVPR'20 \\
    \bottomrule
    \end{tabular}%
  \label{tab:TableLocalFeature}%
\end{table*}%
\autoref{tab:TableLocalFeature} shows the experimental results of the local feature learning methods on CUHK03, Market1501 and DukeMTMC-reID datasets. These results are all experimental results without Re-ranking \citep{Zhong2017Reranking}. In a general view, the experimental performance of the semantic extraction and global-local feature learning methods is significantly higher than that of the methods with predefined stripe segmentation, multi-scale fusion and part attention.

In general, the predefined stripe segmentation method is simple and easy to implement, but it is hard segmentation and requires high image alignment. With the change of real scene camera view and pedestrian pose, the hard segmentation strategy cannot solve the problem of unaligned pedestrians well. The multi-scale fusion method can learn the deeper cues of pedestrian images, but there will be redundancy and conflicting features at different scales. The attention focuses only on the local features of key parts of pedestrians, and easily ignores the distinguishing features of non-focus regions. The semantic extraction method can precisely locate the local features of pedestrians by learning the structural information of pedestrian pose, but it requires the additional computation of pedestrian pose models. The global-local feature learning method can effectively utilize the complementary advantages of global features and local features and is one of the common methods used by researchers to improve model performance.

\subsection{Generative adversarial learning}
\begin{figure*}[tbh]
	\centering
	\includegraphics[width=0.95\linewidth]{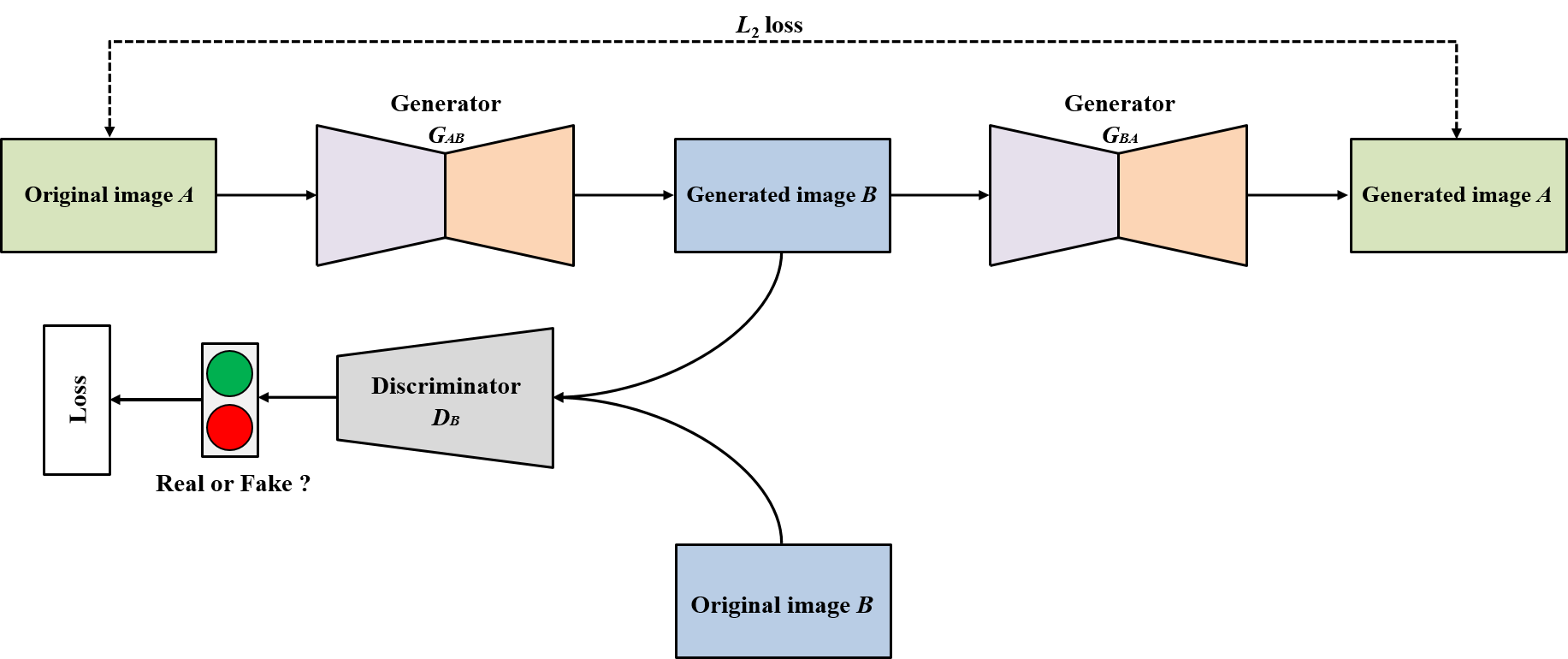}
	\caption{Workflow diagram of GAN for image generation.} 
	\label{fig:3-3-GAN}
\end{figure*}

In 2014, Goodfellow et al. \citep{Goodfellow2014GenerativeAdversarial} first proposed generative adversarial networks (GAN) and rapidly developed in recent years. A large number of variants and applications of GAN emerged \citep{Wei2018PersonTransfer, 2017Image-ImageDeng, 2018CameraZhong, Liu2018PoseTransferrable, Zheng2019JointDiscriminative, 2018FD-GANGe, 2019LearningResolutionChen, Zhu2017UnpairedImage, Choi2018Choi}. Image generation, as one of the significant applications of GAN, was widely used in the field of person Re-ID. \autoref{fig:3-3-GAN} shows the workflow diagram of GAN used to generate the image. In the training phase, the generator ${G}_{AB}$ converts image $A$ into image $B$ with random noise, the generator ${G}_{BA}$ converts image $B$ into image $A$, and the discriminator ${D}_{B}$ determines whether the generated image $B$ approximates the original image $B$ style (Real or Fake). The generator and discriminator keep adversarial until convergence by minimizing the discriminator loss and ${L}_{2}$ loss \citep{Luo2019SurveyDeep}.

Some researchers used GAN to transform the style of images or unify different image styles to mitigate image style differences between various datasets or within the same dataset \citep{Wei2018PersonTransfer,2017Image-ImageDeng,Liu2019AdaptiveTransfer,2020InstanceChen,2018GeneralizingZhong,2018M2MLiang,2018CameraZhong,2020UnityLiu,2019LearningReduceWang,2018DomainBak}. Some works used GAN to synthesize pedestrian images with a different pose, appearance, lighting, and resolution for expanding the dataset to improve the generalization ability of the model \citep{2017UnlabeledZheng,Yan2019Multi,2019PersonRerankingDai,2020Multi-ScaleYang,Liu2018PoseTransferrable,2018Pose-NormalizedQian,2019ProgressiveZhu,2018DisentangledMa,Zheng2019JointDiscriminative,Chen2021JointGenerative}. Some researchers also used GAN to learn features that are not noise-related but identity-related to improve the accuracy of feature matching \citep{2020RealHuang,2018FD-GANGe,2019LearningResolutionChen,2019RecoverLi}. These methods can alleviate the small number of training samples, resolution, illumination, view, and pose variation. Based on the characteristics and application scenarios of GAN, we classify the generative adversarial learning-based person Re-ID methods into three categories: image-image style transfer, data enhancement, and invariant feature learning. For image-image style transfer methods, GAN learned the background, resolution, lighting and other features of an image and transferred these features to other images to give other images a different style. For data enhancement methods, the diversity of samples that can be generated by GAN to expand the dataset was used to reinforce the final feature representation. For invariant feature learning, GAN was used for disentangled representation learning, which can learn identity-related but noise-independent features (e.g., pose, lighting, resolution, etc.).

\subsubsection{Image-image style transfer}
Domain gaps usually exist between different datasets in the person Re-ID task \citep{Wei2018PersonTransfer}. When training and testing on various datasets separately, the performance of the model is severely degraded, which hinders the effective generalization of the model to new test sets \citep{Fan2018Unsupervised,Torralba2011Patch}. A common strategy to solve such as domain gap is to use GAN to perform style transformation across data domains. Since CycleGAN \citep{Zhu2017UnpairedImage} implemented conversion of any two image styles, researchers considered improving on this to achieve adaptive pedestrian style conversion between different datasets to reduce or eliminate domain aberrations. Inspired by CycleGAN, Wei et al. \citep{Wei2018PersonTransfer} proposed a person transfer generative adversarial network (PTGAN) to transfer the pedestrians in the source domain to the target dataset while preserving the identity of the pedestrians in the source domain so that the pedestrians in the source domain have the background and lighting styles of the target domain. The pedestrians in the source domain are transferred to the target dataset so that the pedestrians in the source domain have the background and lighting patterns of the target domain. 

Deng et al. \citep{2017Image-ImageDeng} used twin networks and CycleGAN to form a similarity preserving generative adversarial network (SPGAN) to migrate labeled pedestrians from the source domain to the target domain in an unsupervised manner. Liu et al. \citep{Liu2019AdaptiveTransfer} proposed an adaptive transfer network (ATNet). ATNet used three CycleGANs to implement the style of camera view, lighting, resolution and adaptively assign weights to each CycleGAN according to the degree of influence of different factors. 
Zhong et al. \citep{2018GeneralizingZhong} proposed hetero-homogeneous learning (HHL) method that considers not only the domain differences between various datasets but also the effect of style differences of cameras within the target domain on the cross-domain adaptation person Re-ID performance.

Zhong et al. \citep{2018CameraZhong} introduced camera style(CamStyle) to solve the problem of style variation among different cameras within the same dataset. CamStyle used CycleGAN to migrate the labeled training data to various cameras so that the synthesized samples had the styles of different cameras while retaining the pedestrian labels. In addition, CamStyle can also smooth out style differences between various cameras in the same dataset.

\subsubsection{Data enhancement}
Unlike style transformation that uses GAN to reduce domain gaps, the data augmentation-based methods start with the training of the model and improve the generalization ability of the model by increasing the diversity of the training data. Zheng et al. \citep{2017UnlabeledZheng} were the first to use deep convolutional generative adversarial networks (DCGAN) \citep{Liu2019DeepAdversarial} to generate sample data. Huang et al. \citep{Yan2019Multi} proposed a multi-pseudo regularized label (MpRL), which assigned an appropriate virtual label to each generated sample to establish the correspondence between the real image and the generated image. MpRL effectively distinguished various generated data and achieved good recognition results on datasets such as Market-1501, DukeMTMC-Reid and CUHK03.

Liu et al. \citep{Liu2018PoseTransferrable} introduced pedestrian pose information to assist GAN in generating samples. The GAN was used to generate sample images with both the pose structure of pedestrians in MARS data and the appearance of pedestrians in the existing dataset. Qian et al. \citep{2018Pose-NormalizedQian} used pose normalization GAN (PN-GAN) to generate pedestrian images with uniform body pose. To alleviate the problem that the pose of pedestrian images generated by earlier methods is prone to large deviations, Zhu et al. \citep{2019ProgressiveZhu} trained a discriminator using a multilayer cascaded attention network. The discriminator can efficiently optimize the pose transformation of pedestrians using pose and appearance features so that the generated pedestrian images have the better pose and appearance consistency with the input images.

\begin{table*}[htbp]
  \centering
  \caption{Comparison of experimental results of different GAN-based methods on CUHK03, Market1501 and DukeMTMC-reID datasets.}
    \begin{tabular}{cp{9.125em}p{2.625em}lp{2.69em}p{2.75em}p{4em}p{4em}p{3.565em}}
    \toprule
    \multicolumn{1}{c}{\multirow{2}[4]{*}{Categories}} & \multirow{2}[4]{*}{Methods} & \multicolumn{2}{p{5em}}{CUHK03} & \multicolumn{2}{p{5.44em}}{Market-1501} & \multicolumn{2}{p{8em}}{DukeMTMC-reID} & \multirow{2}[4]{*}{Reference} \\
\cmidrule{3-8}          & \multicolumn{1}{c}{} & mAP   & \multicolumn{1}{p{2.375em}}{R-1} & mAP   & R-1   & mAP   & R-1   & \multicolumn{1}{c}{} \\
    \midrule
    \multicolumn{1}{c}{\multirow{2}[2]{*}{Image-image style transfer}} & IDE\citep{Zheng2016Person}+CameraStyle\citep{2018CameraZhong} &  -    & \multicolumn{1}{p{2.375em}}{ -} & \multicolumn{1}{l}{68.7} & \multicolumn{1}{l}{88.1} & \multicolumn{1}{l}{57.6} & \multicolumn{1}{l}{78.3} & CVPR'18 \\
          & IDE\citep{Zheng2016Person}+UnityStyle\citep{2020UnityLiu} &  -    & \multicolumn{1}{p{2.375em}}{ -} & \multicolumn{1}{l}{\textbf{89.3}} & \multicolumn{1}{l}{\underline{93.2}} & \multicolumn{1}{l}{\underline{65.2}} & \multicolumn{1}{l}{\underline{82.1}} & CVPR'20 \\
    \midrule
    \multicolumn{1}{c}{\multirow{5}[2]{*}{Data enhancement}} & LSRO\citep{2017UnlabeledZheng} & \multicolumn{1}{l}{87.4} & 84.6  & \multicolumn{1}{l}{66.1} & \multicolumn{1}{l}{84.0} & \multicolumn{1}{l}{47.1} & \multicolumn{1}{l}{67.7} & ICCV'17 \\
          & PNGAN\citep{2018Pose-NormalizedQian} &  -    & 79.8 & \multicolumn{1}{l}{72.6} & \multicolumn{1}{l}{89.4} & \multicolumn{1}{l}{53.2} & \multicolumn{1}{l}{73.6} & ECCV'18 \\
          & PT\citep{Liu2018PoseTransferrable} & \multicolumn{1}{l}{42.0} & 45.1  & \multicolumn{1}{l}{68.9} & \multicolumn{1}{l}{87.6} & \multicolumn{1}{l}{56.9} & \multicolumn{1}{l}{78.5} & CVPR'18 \\
          & MpRL\citep{Yan2019Multi} & \multicolumn{1}{l}{\underline{87.5}} & 85.4 & \multicolumn{1}{l}{67.5} & \multicolumn{1}{l}{85.8} & \multicolumn{1}{l}{58.6} & \multicolumn{1}{l}{78.8} & TIP'19 \\
          & DG-Net\citep{Zheng2019JointDiscriminative} &  -    & \multicolumn{1}{p{2.375em}}{ -} & \multicolumn{1}{l}{\underline{86.0}} & \multicolumn{1}{l}{\textbf{94.8}} & \multicolumn{1}{l}{\textbf{74.8}} & \multicolumn{1}{l}{\textbf{86.6}} & CVPR'19 \\
    \midrule
    \multicolumn{1}{c}{\multirow{4}[2]{*}{Invariant feature learning}} & FD-GAN\citep{2018FD-GANGe} & \multicolumn{1}{l}{\textbf{91.3}} & \textbf{92.6} & \multicolumn{1}{l}{77.7} & \multicolumn{1}{l}{90.5} & \multicolumn{1}{l}{64.5} & \multicolumn{1}{l}{80.0} & NIPS'18 \\
          & RAIN\citep{2019LearningResolutionChen} &  -    & 78.9  &  -    &  -    &  -    &  -    & AAAI'19 \\
          & CAD-Net\citep{2019RecoverLi} &  -    & 82.1  &  -    & \multicolumn{1}{l}{83.7} &  -    & \multicolumn{1}{l}{75.6} & ICCV'19 \\
          & DI-REID\citep{2020RealHuang} &  -    & \underline{85.7}  &  -    &  -    &  -    &  -    & CVPR'20 \\
    \bottomrule
    \end{tabular}%
  \label{tab:TableGAN}%
\end{table*}%

\subsubsection{Invariant feature learning}
GAN can be used for feature learning in addition to synthesizing images. Generally, the person Re-ID task in the real-world consists of high-level and low-level vision variations \citep{2020RealHuang}. The former mainly includes changes in pedestrian occlusion, pose, and camera view, and the latter mainly includes changes in resolution, illumination, and weather. The images obtained from low-level vision changes are usually called degraded images. These vision changes may lead to the loss of discriminative feature information, which may cause feature mismatch and significantly degrade the retrieval performance \citep{2019ResolutionMao}. 

For pose changes in high-level vision, Ge et al. \citep{2018FD-GANGe} proposed feature distilling generative adversarial network (FD-GAN) to learn features related to pedestrian identity instead of pose for pedestrians. The method requires no additional computational cost or auxiliary attitude information and has advanced experimental results on the Market-1501, CUHK03 and DukeMTMC-reID.

Several researchers considered using GAN to learn common invariant features of low- and high-resolution pedestrian images. Chen et al. \citep{2019LearningResolutionChen} proposed an end-to-end resolution adaptation and re-identification network (RAIN) that learned and aligned invariant features of pedestrian images of different resolutions by adding adversarial losses to low- and high-resolution image features. Li et al. \citep{2019RecoverLi} proposed adversarial learning strategies for cross resolution, which not only learned invariant features of pedestrian images of different resolutions but also recovered the lost fine-grained detail information of low-resolution images using super-resolution (SR).

\begin{table}[htbp]
  \centering
  \caption{Comparison of experimental results of different GAN-based cross-domain methods.}
    \begin{tabular}{p{7.15em}llllp{3.29em}}
    \toprule
    \multirow{2}[4]{*}{Methods} & \multicolumn{2}{p{3.69em}}{D → M} & \multicolumn{2}{p{3.505em}}{M→ D} & \multirow{2}[4]{*}{Reference} \\
\cmidrule{2-5}    \multicolumn{1}{c}{} & \multicolumn{1}{p{1.875em}}{mAP} & \multicolumn{1}{p{1.815em}}{R-1} & \multicolumn{1}{p{1.69em}}{mAP} & \multicolumn{1}{p{1.815em}}{R-1} & \multicolumn{1}{c}{} \\
    \midrule
    LOMO\citep{2015PersonLiao} & 8.0  & 27.2  & 4.8   & 12.3  & CVPR'15 \\
    Bow\citep{Zheng2015Scalable} & 14.8  & 35.8  & 8.3   & 17.1  & ICCV'15 \\
    \midrule
    UMDL\citep{2016UnsupervisedPeng} & 12.4  & 34.5  & 7.3   & 18.5  & CVPR'16 \\
    CAMEL\citep{Wang2016CrossScenario} & 26.3  & 54.5  & \multicolumn{1}{p{1.69em}}{ -} & \multicolumn{1}{p{1.815em}}{ -} & ICCV'17 \\
    PUL\citep{Zha2020AdversarialAttribute} & 20.5  & 45.5  & 16.4  & 30.0    & MM'18 \\
    \midrule
    CycleGAN\citep{Zhu2017UnpairedImage} & 19.1  & 45.6  & 19.6  & 38.1  & CVPR'17 \\
    PTGAN\citep{Wei2018PersonTransfer} & \multicolumn{1}{p{1.875em}}{ -} & 38.6  & 27.4  & \multicolumn{1}{p{1.815em}}{ -} & CVPR'18 \\
    SPGAN\citep{2017Image-ImageDeng} & 22.8  & 51.5  & 22.3  & 41.4  & CVPR'18 \\
    HHL\citep{2018GeneralizingZhong} & 31.4 & 62.2 & 27.2  & 46.9  & ECCV'18 \\
    ATNet\citep{Liu2019AdaptiveTransfer} & 25.6  & 55.7  & 24.9  & 45.1  & CVPR'19 \\
    CR-GAN\citep{2020InstanceChen} & 29.6  & 59.6  & 30.0 & 52.2 & ICCV'19 \\
    
    DG-Net++\citep{Zou2020Joint} & \underline{61.7}  & \underline{82.1}  & \underline{63.8} & \underline{78.9} & ECCV'20 \\
    GCL\citep{Chen2021JointGenerative} & \textbf{75.4}  & \textbf{90.5}  & \textbf{67.6} & \textbf{81.9} & CVPR'21 \\
    \bottomrule
    \end{tabular}%
  \label{tab:TableGANCrossDomain}%
\end{table}%

\subsubsection{Comparison and discussion}
\autoref{tab:TableGAN} shows the experimental results of GAN-based methods on CUHK03, Market1501 and DukeMTMC-reID datasets. These results are all experimental results without Re-ranking. Both IDE \citep{Zheng2016Person} + CameraStyle \citep{2018CameraZhong} and IDE \citep{Zheng2016Person} + UnityGAN \citep{2020UnityLiu} methods use GAN to generate pedestrian images with different camera styles within the same dataset with great experimental performance obtained in the Market-1501 and DukeMTMC-reID datasets. FD-GAN \citep{2018FD-GANGe} has the highest performance in CUKH03, and it learns features related to pedestrian identity and poses independently, effectively reducing the influence of pose on the accuracy of person Re-ID.

We use the Market-1501 (M) and DukeMTMC-reID (D) datasets as the source and target domains, respectively. \autoref{tab:TableGANCrossDomain} compares traditional manual feature extraction methods (LOMO \citep{2015PersonLiao} and Bow \citep{Zheng2015Scalable}), traditional unsupervised methods (UMDL \citep{2016UnsupervisedPeng}, CAMEL \citep{Wang2016CrossScenario} and PUL \citep{Zha2020AdversarialAttribute}), and GAN-based cross-domain style transfer methods (CycleGAN(base) \citep{Zhu2017UnpairedImage}, PTGAN \citep{2017Image-ImageDeng}, SPGAN \citep{2017Image-ImageDeng}, HHL \citep{2018GeneralizingZhong}, ATNet \citep{Liu2019AdaptiveTransfer} and CR-GAN \citep{2020InstanceChen}). From the experimental results, the cross-domain style transformation method is significantly better than the traditional unsupervised learning and manual feature learning methods.

In general, the method of image-to-image style transformation smooths the style variation of pedestrian images in different domains. Such methods can obtain a large number of automatically labeled synthetic images with the target domain style, which can be used together with the original images to enhance the training set and mitigate the domain gaps between different datasets. The problem with these methods is that the synthetic images contain noise, which may conflict with the source domain images when used for model training and affect the learning of discriminative features by the model. The method of generating diverse pedestrian images using GAN alleviates the problem of insufficient available training data to a certain extent. The method of image synthesis without auxiliary information guidance cannot generate high-quality images with sufficient distinguishing information. Auxiliary information-guided image synthesis methods require complex network structures to learn various pedestrian poses, which adds additional training costs. Invariant feature learning methods can alleviate the problem of unaligned pedestrian features and improve the accuracy of 
person Re-ID by learning features related to pedestrian identity but not to pose, resolution and illumination.

\subsection{Sequence feature learning}
There have been many researchers who have used the rich information contained in video sequences for person Re-ID. These sequence feature learning-based methods take short videos as input and use both spatial and temporal complementary cues, which can alleviate the limitations of appearance-based features. Most of these methods use optical flow information \citep{Chung2017StreamSiamese,Liu2018VideoBased,Chen2018VideoPerson,McLaughlin2016RecurrentConvolutional,Xu2017JointlyAttentive}, 3-dimensional convolutional neural networks (3DCNNs) \citep{Liao2019VideoBased,Li2019MultiScale}, recurrent neural networks(RNN) or long short term memory(LSTM) \citep{Chen2018VideoPerson,McLaughlin2016RecurrentConvolutional,Zhou2017ForestTrees,Yan2016PersonIdentification}, spatial-temporal attention \citep{Fu2019SpatialTemporal,Hou2019VrstcOcclusion,Xu2017JointlyAttentive,Li2018DiversityRegularized,Zhang2020MultiGranularity,Li2019GlobalLocal,Hou2021BiCnet} or graph convolutional networks (GCN) \citep{Wu2020AdaptiveGraph,Yan2020LearningMulti,Yang2020SpatialTemporal,Liu2021Spatial} to model the spatial-temporal information of video sequences. These methods can mitigate occlusions, resolution changes, illumination changes, view and pose variations, etc.

\subsubsection{Optical flow}
The optical flow method uses the change of pixels in the video sequence in the time domain and the correlation of the spatial-temporal context of adjacent frames to obtain the correspondence between the previous frame and the current frame. This method can obtain the motion information of the target between adjacent frames. Chung et al. \citep{Chung2017StreamSiamese} proposed a dual-stream convolutional neural network (DSCNN), where each stream is a siamese network. DSCNN can model both RGB images and optical flow, learning spatial and temporal information separately, allowing each siamese network to extract the best feature representation. Liu et al. \citep{Liu2018VideoBased} proposed an accumulative context network (AMOC), which consists of two input sequences, feeding the original RGB image and the optical flow image containing motion information, respectively. AMOC was used to improve the accuracy of person Re-ID by learning the discriminative cumulative motion context information of video sequences. The optical flow method was often used in combination with other methods such as McLaughlin et al. \citep{McLaughlin2016RecurrentConvolutional} who used optical flow information and RGB colors of images to capture motion and appearance information, combined with RNN to extract complete pedestrian appearance features of video sequences.

\subsubsection{3D convolutional neural network}
Three-dimensional convolutional neural networks (3DC-NN) are capable of capturing temporal and spatial feature information in videos. Recently, some researchers have applied 3DCNN to video-based person Re-ID with good results. Liao et al. \citep{Liao2019VideoBased} proposed a video person Re-ID method based on a combination of 3DCNN and non-local attention. 3DCNN used 3D convolution on video sequences to extract aggregated representations of spatial and temporal features and used non-local spatial-temporal attention to solve the alignment problem of deformed images. Although 3DCNN exhibited better performance, the stacked 3D convolution led to significant growth of parameters. Too many parameters not only made 3DCNN computationally expensive but also led to difficulties in model training and optimization. This made 3DCNN not readily applicable on video sequence-based person Re-ID, where the training set was commonly small and person ID annotation was expensive. To explore rich temporal cues for person Re-ID while mitigating the shortcomings of existing 3DCNN models, Li et al. \citep{Li2019MultiScale} proposed a dual-stream multiscale 3D convolutional neural network (M3DCNN) for extracting spatial-temporal cues for video-based person Re-ID. M3DCNN was also more efficient and easier to optimize than the existing 3DCNN.

\subsubsection{RNN or LSTM}
RNNs or LSTM can extract temporal features and are often applied in video-based person Re-ID tasks. McLaughlin et al. \citep{McLaughlin2016RecurrentConvolutional} proposed a novel recursive convolutional network(RCN), which used CNN to extract spatial features of video frames and RNNs to extract temporal features of video sequences. Yan et al. \citep{Yan2016PersonIdentification} used a recurrent feature aggregation network based on LSTM, which obtained cumulative discriminative features from the first LSTM node to the deepest LSTM node and effectively alleviated interference caused by occlusion, background clutter and detection failure. Chen et al. \citep{Chen2018VideoPerson} decomposed a video sequence into multiple segments and used LSTM to learn the segments where the probe images are located in temporal and spatial features. This method reduces the variation of identical pedestrians in the sample and facilitates the learning of similarity features. Both types of methods mentioned above process each video frame independently. The features extracted by LSTM are generally affected by the length of the video sequence. The RNN only establishes temporal associations on high-level features and thus cannot capture the temporal cues of local details of the image \citep{Li2019MultiScale}. Therefore, there is still a need to explore a more efficient method for extracting spatial-temporal features.

\subsubsection{Spatial-temporal attention}
The attention mechanism can selectively focus on the useful local information and has good performance in solving the problems of camera view switching, lighting changes and occlusion in person Re-ID tasks. Recently, several researchers have used attentional mechanisms to solve video-based person Re-ID tasks in both temporal and spatial dimensions. To solve the problem of unalignment and occlusion caused by changes in pedestrian body pose and camera view in video sequences, Li et al. \citep{Li2018DiversityRegularized} proposed a spatial-temporal attention model, the core idea of which is to use multiple spatial attention to extract features of key body parts and use temporal attention to compute the combined feature representations extracted by each spatial attention model. This method can better mine the potential distinguishing feature representations in video sequences. Similarly, Fu et al. \citep{Fu2019SpatialTemporal} proposed a spatial-temporal attention framework that can fully utilize the distinguishing features of each pedestrian in both temporal and spatial dimensions through video frame selection, local feature mining, and feature fusion. The approach can well address challenges such as pedestrian pose variation and partial occlusion. Xu et al. \citep{Xu2017JointlyAttentive} proposed a joint temporal and spatial attention pooling network to learn the feature representations of video sequences through the interdependence between video sequences.

\subsubsection{Graph convolutional networks}
In recent years, graph convolutional networks (GCNs) have been widely used for person Re-ID tasks due to their powerful automatic relational modeling capabilities \citep{Kipf2017SemiSupervised}, and a large number of variant networks have emerged \citep{Yang2020SpatialTemporal,Yan2019LearningContext,Shen2018PersonIdentification,Wu2020AdaptiveGraph,Bai2021Unsupervised}. Yang et al. \citep{Yang2020SpatialTemporal} proposed a unified spatial-temporal graph convolutional neural network that modelled video sequences in three dimensions: temporal, spatial and appearance, and to mine more discriminative and robust information. Wu et al. \citep{Wu2020AdaptiveGraph} proposed an adaptive graph representation learning scheme for video person Re-ID using pose alignment connections and feature similarity connections to construct adaptive structure-aware adjacency graphs.

Yan et al. \citep{Yan2019LearningContext} proposed a framework for pedestrian retrieval based on contextual graphical convolutional networks. Since image appearance features are not sufficient to distinguish different people, the authors use contextual information to extend instance-level features to improve the discriminative power of the features and  the robustness of person retrieval. Shen et al. \citep{Shen2018PersonIdentification} proposed a similarity-guided graph neural network that represents pairwise relationships between probe-gallery image pairs (nodes) by creating a graph and using this relationship to enhance the learning of discriminative features. This updated probe-gallery image is used to predict the relational features for accurate similarity estimation.

The last group of \autoref{tab:TableVideoSequence} shows the experimental results of the GCN-based sequence feature learning methods on MARS, DukeMTMC-VideoReID iLIDS-VID and PRID-2011 datasets. From the results in the above table, the experimental performance of the GCN-based methods is significantly better than the other sequence feature learning methods. In particular, CTL \citep{Liu2021Spatial} achieved Rank-1 accuracy of 91.4\% and mAP of 86.7\% on MARS. CTL utilized a CNN backbone and a key-points estimator to extract semantic local features from the human body at multiple granularities as graph nodes. CTL effectively mined comprehensive cues complementary to appearance information to enhance the representation capability.
\subsubsection{Comparison and discussion}
\begin{table*}[htbp]
  \centering
  \caption{Comparison of experimental results of video sequence feature learning-based methods.}
    \begin{tabular}{cp{7.965em}llllp{2.75em}p{2.565em}p{2.75em}p{2.565em}p{3.8em}}
    \toprule
    \multicolumn{1}{c}{\multirow{2}[2]{*}{Categories}} & \multirow{2}[2]{*}{Methods} & \multicolumn{2}{p{4.815em}}{MARS} & \multicolumn{2}{p{4.755em}}{DukeMTMC-VideoReID} & \multicolumn{2}{p{5.315em}}{iLIDS-VID} & \multicolumn{2}{p{5.315em}}{PRID-2011} & \multirow{2}[2]{*}{Reference} \\
          & \multicolumn{1}{c}{} & \multicolumn{1}{p{2.565em}}{mAP} & \multicolumn{1}{p{2.25em}}{R-1} & \multicolumn{1}{p{2.565em}}{mAP} & \multicolumn{1}{p{2.19em}}{R-1} & R-1   & R-5   & R-1   & R-5   & \multicolumn{1}{c}{} \\
    \midrule
    \multicolumn{1}{c}{\multirow{6}[1]{*}{Optical flow}} & RCN\citep{McLaughlin2016RecurrentConvolutional}* & \multicolumn{1}{p{2.565em}}{-} & \multicolumn{1}{p{2.25em}}{-} & \multicolumn{1}{p{2.565em}}{-} & \multicolumn{1}{p{2.19em}}{-} & \multicolumn{1}{l}{58.0} & \multicolumn{1}{l}{84.0} & \multicolumn{1}{l}{70.0} & \multicolumn{1}{l}{90.0} & CVPR'16 \\
          & TSSCNN\citep{Chung2017StreamSiamese} & \multicolumn{1}{p{2.565em}}{-} & \multicolumn{1}{p{2.25em}}{-} & \multicolumn{1}{p{2.565em}}{-} & \multicolumn{1}{p{2.19em}}{-} & \multicolumn{1}{l}{60.0} & \multicolumn{1}{l}{86.0} & \multicolumn{1}{l}{78.0} & \multicolumn{1}{l}{94.0} & ICCV'17 \\
          & ASTPN\citep{Xu2017JointlyAttentive} & \multicolumn{1}{p{2.565em}}{-} & 44.0    & \multicolumn{1}{p{2.565em}}{-} & \multicolumn{1}{p{2.19em}}{-} & \multicolumn{1}{l}{62.0} & \multicolumn{1}{l}{86.0} & \multicolumn{1}{l}{70.0} & \multicolumn{1}{l}{90.0} & ICCV'17 \\
          & AMOC\citep{Liu2018VideoBased} & 52.9  & 68.3  & \multicolumn{1}{p{2.565em}}{-} & \multicolumn{1}{p{2.19em}}{-} & \multicolumn{1}{l}{68.7} & \multicolumn{1}{l}{94.3} & \multicolumn{1}{l}{83.7} & \multicolumn{1}{l}{98.3} & TCSVT'18 \\
          & CSACSE\citep{Chen2018VideoPerson}* & 69.4  & 81.2  & \multicolumn{1}{p{2.565em}}{-} & \multicolumn{1}{p{2.19em}}{-} & \multicolumn{1}{l}{79.8} & \multicolumn{1}{l}{91.8} & \multicolumn{1}{l}{81.2} & \multicolumn{1}{l}{92.1} & CVPR'18 \\
          & CSACSE+OF\citep{Chen2018VideoPerson}* & 76.1  & 86.3  & \multicolumn{1}{p{2.565em}}{-} & \multicolumn{1}{p{2.19em}}{-} & \multicolumn{1}{l}{85.4} & \multicolumn{1}{l}{96.7} & \multicolumn{1}{l}{93.0} & \multicolumn{1}{l}{99.3} & CVPR'18 \\
    \midrule
    \multicolumn{1}{c}{\multirow{2}[0]{*}{3DCNN}} & 3DCNN+NLA\citep{Liao2019VideoBased} & 77.0    & 84.3  & \multicolumn{1}{p{2.565em}}{-} & \multicolumn{1}{p{2.19em}}{-} & \multicolumn{1}{l}{81.3} & -     & \multicolumn{1}{l}{91.2} & -     & ACCV'18 \\
          & M3D\citep{Li2019MultiScale} & 74.1 & 84.4 & \multicolumn{1}{p{2.565em}}{-} & \multicolumn{1}{p{2.19em}}{-} & \multicolumn{1}{l}{74.0} & \multicolumn{1}{l}{94.3} & \multicolumn{1}{l}{94.4} & \multicolumn{1}{l}{\textbf{100.0}} & AAAI'19 \\
    \midrule
    \multicolumn{1}{c}{\multirow{2}[1]{*}{RNN or LSTM}} & RFA\citep{Yan2016PersonIdentification} & \multicolumn{1}{p{2.565em}}{-} & \multicolumn{1}{p{2.25em}}{-} & \multicolumn{1}{p{2.565em}}{-} & \multicolumn{1}{p{2.19em}}{-} & \multicolumn{1}{l}{49.3} & \multicolumn{1}{l}{76.8} & \multicolumn{1}{l}{58.2} & \multicolumn{1}{l}{85.8} & ECCV'16 \\
          & SFT\citep{Zhou2017ForestTrees} & 50.7  & 70.6  & \multicolumn{1}{p{2.565em}}{-} & \multicolumn{1}{p{2.19em}}{-} & \multicolumn{1}{l}{55.2} & \multicolumn{1}{l}{86.5} & \multicolumn{1}{l}{79.4} & \multicolumn{1}{l}{94.4} & CVPR'17 \\
    \midrule
    \multicolumn{1}{c}{\multirow{5}[1]{*}{\shortstack{Spatial-temporal\\attention}}} & DRSA\citep{Li2018DiversityRegularized} & 65.8  & 82.3  & \multicolumn{1}{p{2.565em}}{-} & \multicolumn{1}{p{2.19em}}{-} & \multicolumn{1}{l}{80.2} & -     & \multicolumn{1}{l}{93.2} & -     & CVPR'18 \\
          & GLTR\citep{Li2019GlobalLocal} & 78.5  & 87.0    & 93.7  & 96.3  & \multicolumn{1}{l}{\underline{86.0}} & \multicolumn{1}{l}{\textbf{98.0}} & \multicolumn{1}{l}{\underline{95.5}} & \multicolumn{1}{l}{\textbf{100.0}} & ICCV'19 \\
          & VRSTC\citep{Hou2019VrstcOcclusion} & 82.3  & 88.5  & 93.5  & 95.0    & \multicolumn{1}{l}{83.4} & \multicolumn{1}{l}{95.5} & -     & -     & CVPR'19 \\
          & STA\citep{Fu2019SpatialTemporal} & 80.8  & 86.3  & 94.9  & 96.2  & -     & -     & -     & -     & AAAI'19 \\
          & MG-RAFA\citep{Zhang2020MultiGranularity} & 85.6  & 88.8  & \multicolumn{1}{p{2.565em}}{-} & \multicolumn{1}{p{2.19em}}{-} & \multicolumn{1}{l}{\textbf{88.6}} & \multicolumn{1}{l}{\textbf{98.0}} & \multicolumn{1}{l}{\textbf{95.9}} & \multicolumn{1}{l}{\underline{99.7}} & CVPR'20 \\
          
          & BiCnet-TKS\citep{Hou2021BiCnet} & \underline{86.0}  & \underline{90.2}  & \textbf{96.1}  & 96.3  & 75.1     & 84.6     & -     & -     & CVPR'21 \\
          
    \midrule
    \multicolumn{1}{c}{\multirow{3}[0]{*}{GCN}} & AdaptiveGraph\citep{Wu2020AdaptiveGraph} & 81.9  & 89.5  & 95.4  & \underline{97.0}    & \multicolumn{1}{l}{84.5} & \multicolumn{1}{l}{96.7} & \multicolumn{1}{l}{94.6} & \multicolumn{1}{l}{99.1} & TIP'20 \\
          & MGH\citep{Yan2020LearningMulti} & 85.8  & 90.0 & \multicolumn{1}{p{2.565em}}{-} & \multicolumn{1}{p{2.19em}}{-} & \multicolumn{1}{l}{85.6} & \multicolumn{1}{l}{\underline{97.1}} & \multicolumn{1}{l}{94.8} & \multicolumn{1}{l}{99.3} & CVPR'20 \\
          & STGCN\citep{Yang2020SpatialTemporal} & 83.7 & 89.9  & \underline{95.7} & \textbf{97.3} & -     & -     & -     & -     & CVPR'20 \\
          
          & CTL\citep{Liu2021Spatial} & \textbf{86.7} & \textbf{91.4}  & - & - & -     & 89.7     & -     & -     & CVPR'21 \\
          
    \bottomrule
    \end{tabular}%
  \label{tab:TableVideoSequence}%
\end{table*}%

\autoref{tab:TableVideoSequence} shows the experimental results of sequence feature learning-based methods on MARS, DukeMTMC-VideoReID, iLIDS-VID and PRID-2011 datasets. Some researchers used GCN to model the Spatio-temporal relationships of video sequences and achieved good results. Compared with optical flow, 3DCNN, and RNN or LSTM methods, spatial-temporal attention-based methods can obtain better experimental performance. AdaptiveGraph \citep{Wu2020AdaptiveGraph}, MGH \citep{Yan2020LearningMulti} and STGCN \citep{Yang2020SpatialTemporal} were able to obtain high experimental results on the above datasets, with some improvement in accuracy compared to the previous types of methods.

The core idea of the sequence feature learning-based methods is to fuse more spatial-temporal information from multiple dimensions to mitigate the effects of a range of problems such as occlusion, illumination, and viewpoint changes in the person Re-ID tasks. Although Optical flow can provide contextual information of video sequence frames,  it only represents the local dynamics of adjacent images, which may introduce noise due to spatial misalignment. The process of computing optical flow is time-consuming. 3DCNN can capture both temporal and spatial feature information in video sequences. Although 3DCNN can achieve better performance, these methods are computationally time-consuming and hard to optimize. RNN or LSTM can extract temporal features of video sequences and has been popular among researchers for some time. In the task of person Re-ID, the ability of RNN or LSTM have limited features for temporal information extraction and suffer from difficulties in model training due to the complex network structure \citep{Gao2018RevisitingTemporal}. Although the introduction of temporal attention and spatial attention can alleviate the problem of switching between different camera views in a row, lighting changes and occlusion, the accuracy of person Re-ID is affected because of the temporal relationship between body parts in different frames is not fully considered \citep{Yang2020SpatialTemporal}. Most of the current person Re-ID studies are image-based, and a large number of methods and datasets closer to the real world have emerged. Compared to image-based methods, sequence feature learning-based methods still hold great research promise.

\section{Conclusion and future directions}\label{sec:ConclusionAndFutureDirections} 
This paper presents a comprehensive survey with in-depth discussion for deep learning-based person Re-ID methods in recent years. Firstly, we summarize the main contributions of several recently published person Re-ID surveys and discuss the common datasets used for person Re-ID benchmarks. Secondly, we comprehensively review the current deep learning-based methods, these methods are classified into four major categories according to metric learning and representation learning, including deep metric learning, local feature learning, generative adversarial learning and sequence feature learning. 
We subdivide the above four categories according to their methodologies and motivations, analyzing and discussing the advantages and limitations of each subcategory of the method. This classification is more suitable for researchers to explore these methods from their practical needs.

Although existing deep learning-based methods have achieved good results in person Re-ID tasks, they still face many challenges. Most of the datasets currently applied for person Re-ID training are processed visible images or videos, but real-world data often exhibit a combination of multiple modalities. Although semi-supervised and unsupervised methods can alleviate the problem of high labeling costs, they still do not perform as well as supervised methods. There are domain differences in pedestrian images captured by different cameras, and models trained on one dataset can experience severe performance degradation when tested on another dataset. Because people may change their clothes or different people may wear very similar clothes, the appearance features of pedestrians will become unreliable for person Re-ID. In addition, how to improve the speed and accuracy of model retrieval is critical for real-world model deployment. The increase of privacy scenarios fundamentally limits the traditional centralized person Re-ID methods. The detection and re-identification modules of most person Re-ID systems are separated from each other, making it difficult to expand to real-world applications. 

In summary, there are still many challenges to be explored and researched in deep learning-based person Re-ID methods. The following subsections present potential solutions to address the above existing challenges, as well as prospects of future research directions.

\textbf{(1) Cross-modal person Re-ID.} Most existing Re-ID methods evaluate their performance on publicly available datasets, which are obtained based on image or video processing. However, the acquisition of real-world data is diverse and the data may appear as a combination of different modalities (visible, infrared, depth map and text descriptions, etc.). For example, in the absence of sufficient visibility information like images or videos, text descriptions can provide unique attributes aiding information for person Re-ID. Several research works\citep{Zha2020AdversarialAttribute,Niu2020Improving}  learned discriminative cross-modal visual-textual features for better similarity evaluation in description-based person Re-ID. Because it was difficult for visible-light cameras to capture valid appearance information in dark environments, some researchers \citep{Choi2020Choi,Ye2020Dynamic,Wang2020CrossModality,Li2020Infrared,Wu2021Discover} used thermal infrared images to learn rich visual representations  for cross-modality matching. Existing works mainly focus on alleviating the modality discrepancy by aligning the distributions of features from different modalities. Meanwhile, how to combine various modal complementary information is also worth studying in the future.

\textbf{(2) High-performance semi-supervised and unsupervised person Re-ID.} Because it was expensive to annotate person images across multiple cameras, some researchers \citep{Yu2019UnsupervisedPerson,Lin2020UnsupervisedPerson,Wang2020UnsupervisedPerson,Yang2019PatchBased,Xin2019SemiSupervised,Qi2020ProgressiveCross} focused on semi-supervised and unsupervised methods for person Re-ID. These methods aimed to learn discriminative features from unlabeled or minimally labeled images of people. Compared to supervised learning, semi-supervised and unsupervised methods alleviated the need for expensive data annotation and showed great potential to facilitate person Re-ID to practical applications. Some semi-supervised person Re-ID methods \citep{Tang2019Unsupervised,Wu2020Tracklet} made use of image clustering or tracklet clustering in the target domain to adapt the model to the new domain. Some unsupervised person Re-ID methods \citep{Yu2019UnsupervisedPerson,Lin2020UnsupervisedPerson,Wang2020UnsupervisedPerson} used soft labels or multi-labels to learn discriminative embedding features. Although lacking realistic label learning discriminative features, the performance of person Re-ID methods in semi-supervised and unsupervised scenarios was still inferior to that of supervised methods, they still maintained significant research value and significance in improving the generalization ability of the model \citep{Luo2019SurveyDeep}. In future research, better clustering or label assignment strategies should be considered to improve the performance of person Re-ID.

\textbf{(3) Domain adaptation person Re-ID.} The background, resolution, and illumination environment of different cameras in the real world vary greatly, which interfered with the learning of distinguishing features of pedestrians and affected the performance of person Re-ID. Some researchers \citep{Wei2018PersonTransfer,2017Image-ImageDeng,2020InstanceChen,Liu2019AdaptiveTransfer} transferred images with identity labels from the source to the target domain to learn discriminative models, but they largely ignored the unlabeled samples and the substantial sample distributions in target domains. Some researchers \citep{Fan2018Unsupervised,Fu2019Self,Zhai2020ADCluster,Zhang2019SelfTraining,Ye2017Dynamic} used clustering or graph matching methods to predict the pseudo-labels in the target domain for discriminative model learning, but they still faced the challenge of accurately predicting hard samples labels. Domain adaption is crucial for person Re-ID models learned in unknown domains. Therefore, it remains one of the important research directions for the future.

\textbf{(4) Person Re-ID in the 3D space.} In the real world, the spatial location of cameras is uncertain and a new camera may be temporarily inserted into an existing camera network. Considering people may change their clothes or different people may wear very similar clothes, pedestrian appearance features will become unreliable for Re-ID \citep{Chen2021Learning}. The 3D structure does not rely on the appearance information of 2D images can effectively alleviate this limitation. However, the acquisition of 3D point cloud data of pedestrians requires additional auxiliary models. Some researchers \citep{Chen2021Learning,Zheng2020Parameter} extracted 3D shape embeddings directly from 2D images, obtained more robust structural and appearance information by aligning 2D and 3D local features, or planed 3D models back to 2D images for representation learning in 2D space for data augmentation purposes. Although the above studies achieved good experimental results, the 2D data space inherently limited the model to understand the 3D geometric information of people. Therefore, further exploration of person Re-ID methods in 3D space is still an important research direction in the future.

\textbf{(5) Fast person Re-ID.} Most current methods to person Re-ID focus mainly on prior knowledge or designing complex network architectures to learn robust identity invariant feature representations. These methods use complex network models to extract high-dimensional features to improve model performance. However, the above methods use Euclidean distance to calculate the similarity of features and obtain the rank list by fast sorting, which will increase with the retrieval time as the size of the gallery library increases. This retrieval method will be very time-consuming, making the model unsuitable for real-world applications. Therefore, some researchers \citep{Chen2017Fast, Wang2020Faster, Zhao2021Salience} considered introducing hashing to improve the retrieval speed. Faster person Re-ID retrieval from coarse-to-fine (CtF) \citep{Wang2020Faster} can be achieved by supplementing long and short hash codes to get faster and better accuracy. Zhao et al. \citep{Zhao2021Salience} proposed saliency-guided iterative asymmetric mutual hashing (SIAMH) to achieve high-quality hash code generation and fast feature extraction. However, how to design a specific retrieval strategy to reduce the information redundancy among models and improve retrieval speed and accuracy still needs further research.

\textbf{(6) Decentralised learning person Re-ID.}
Most of the existing person Re-ID methods use a centralized learning paradigm, which requires collecting all training data from different camera views or domains for centralized training. Although these supervised or unsupervised methods have made significant progress, centralized person Re-ID learning ignores images of people that contain large amounts of personal and private information that may not be allowed to be shared into a central data set. As privacy scenarios increase, it can fundamentally limit the centralized learning person Re-ID methods in the real world. Several recent works \citep{Wu2020Decentralised,Sun2021DecentralisedPerson,Zhuang2020Performance,Bedogni2021Modelling} attempted to address the above problem through decentralised learning. These methods either built a globally generalised model server through federated learning, which did not require access to local training data and shared of cross-domain data, or selectively performed knowledge aggregation to optimize the trade-off between model personalisation and generalization in decentralised person Re-ID. In future work, how to ensure understanding cross-domain data heterogeneity while learning a global generalised model remains challenging.

\textbf{(7) End-to-end person Re-ID system.} As is shown in \autoref{fig:1-1-Re-ID-Pipeline} in the introduction of this paper, person detection and re-identification in most current person Re-ID systems are two independent modules. The person Re-ID task uses the correct pedestrian already detected by default, but some practical open-world applications require end-to-end person search from the raw images or videos \citep{Xiao2017Joint}. The two-stage end-to-end person re-identification framework was one of the most common Re-ID systems that systematically evaluated the advantages and limitations of combining different detectors and Re-ID models \citep{Zheng2017Wild}. Munjal et al. \citep{Munjal2019Query} proposed a query-guided end-to-end person search network (QEEPS) to join person detection and re-identification. In addition, the end-to-end person Re-ID was also widely used in multi-target multi-camera tracking (MTMC tracking) \citep{ Ristani2018Multitarget, Tang2017Multiple, Hou2019Locality}. Person Re-ID algorithms rely not only on accurate person detection algorithms but also on detected unlabeled pedestrians, which remains a current challenge. Therefore, how to effectively combine person detection and re-identification to design an end-to-end person Re-ID system is also a direction that researchers need to pay attention to in the future.








\section*{Declaration of Competing Interest}{The authors declare that they have no known competing financial interests or personal relationships that could have appeared to influence the work reported in this paper.}

\section*{Acknowledgement}{The authors wish to thank Jizhuo Li, Xiao Pang, Jiamin Zhu, Fuqiu Chen, Yi Zhou and Cheng Zhang. This work was supported by the National Key Research and Development Project of China (No. JG2018190), and in part by the National Natural Science Foundation of China (No. 61872256).}

\bibliographystyle{model1-num-names}

\bibliography{cas-refs}



\end{document}